\documentclass{article} 
\usepackage{shmiclr2023_conference,times}
\usepackage[font=small,labelfont=bf]{caption}

\usepackage{amsmath,amsfonts,bm}

\def\eqref#1{equation~\ref{#1}}

\def\1{\bm{1}}

\DeclareMathAlphabet{\mathsfit}{\encodingdefault}{\sfdefault}{m}{sl}
\SetMathAlphabet{\mathsfit}{bold}{\encodingdefault}{\sfdefault}{bx}{n}

\usepackage{booktabs}
\usepackage{hyperref}
\usepackage{url}
\definecolor{cornflowerblue}{rgb}{0.39, 0.58, 0.93}
\hypersetup{
    colorlinks=true,
    linkcolor=cornflowerblue,
    filecolor=magenta,      
    urlcolor=teal,
    citecolor=cornflowerblue,
    pdftitle={Cramming: Training a Language Model on a Single GPU in One Day},
    pdfpagemode=FullScreen,
    }
    
\newcommand{\bitet}[1]{{\hypersetup{citecolor=gray}\citet{#1}}}
\newcommand{\bitep}[1]{{\hypersetup{citecolor=gray}\citep{#1}}}
\usepackage{multirow}

\usepackage{xurl}

\usepackage{graphicx}
\usepackage[nameinlink,capitalise,noabbrev]{cleveref} 
\usepackage{amssymb}
\usepackage{siunitx}
\usepackage{wrapfig}

\title{\centering Cramming: Training a Language Model on a Single GPU in One Day}

\author{
Jonas Geiping \\
University of Maryland, College Park\\
\texttt{jgeiping@umd.edu}
\And 
Tom Goldstein\\
University of Maryland, College Park\\
\texttt{tomg@umd.edu}
}

\iclrfinalcopy 
\begin{document}

\maketitle

\begin{abstract}

Recent trends in language modeling have focused on increasing performance through scaling, and have resulted in an environment where training language models is out of reach for most researchers and practitioners.  While most in the community are asking how to push the limits of extreme computation, we ask the opposite question:  
How far can we get with a single GPU in just one day? 

\looseness -1 We investigate the downstream performance achievable with a transformer-based language model trained completely from scratch with masked language modeling for a \textit{single} day on a \textit{single consumer} GPU.
Aside from re-analyzing nearly all components of the pretraining pipeline for this scenario and providing a modified pipeline with performance close to BERT, we investigate why scaling down is hard, and which modifications actually improve performance in this scenario. We provide evidence that even in this constrained setting, performance closely follows scaling laws observed in large-compute settings. Through the lens of scaling laws, we categorize a range of recent improvements to training and architecture and discuss their merit and practical applicability (or lack thereof) for the limited compute setting.
\end{abstract}

\section{Scaling Up and Scaling Down}

Large-scale training of machine learning models with transformer architectures has lead to ground-breaking improvements in many sub-fields of natural language processing including language understanding and natural language generation \citep{vaswani_attention_2017,dosovitskiy_image_2021,radford_language_2019}. The nowadays accepted (but historically surprising) key behavior of these systems is that they reliably \textit{scale} -- they continuously improve in performance when the number of model parameters and amount of data grow.  These increases in performance are well-described by various power laws as studied by \citet{kaplan_scaling_2020}. This sets up a dominant paradigm in which scaling is the key to performance improvement \citep{sutton_bitter_2019}.


The power of scale has set off a race to produce extremely large models, which in turn has created an environment where few researchers or practitioners feel that they are capable of training a language model. The original BERT model \citet{devlin_bert_2019}, which became a cornerstone transformer for many practical applications in natural language understanding, already required a significant amount of computation to train.  Yet, the reproduction and improvements in \citet{liu_roberta_2019} further increased its performance by cranking up the level of computation by orders of magnitude.  As these pre-trained checkpoints became popular for a range of downstream applications \citep{wolf_huggingfaces_2020}, the competition for the largest language model became a focal point for industrial labs.
This led to training runs that improved the performance of pretrained language models at the expense of computation at the zettaFLOP scale \citep{raffel_exploring_2020,yang_xlnet_2020,zaheer_big_2021} and later at the extremely large yottaFLOP scale \citep{brown_language_2020,black_gpt-neox-20b_2022,chowdhery_palm_2022,rae_scaling_2022}.

Our goal is to turn this trend on its head and investigate how to best {\em scale down} language model training and what trade-offs emerge when doing so:  {\em What downstream performance can be achieved by a modest researcher when training from scratch with a single GPU for a single day?}
\looseness -1 The ability to train a language model to the performance level of BERT with such modest resources has several interesting implications. For one, if scaled-down model pretraining is a viable analogue of large-compute pretraining, then this opens up a host of further academic investigations that are currently hard to realize for large-scale models. For example, research questions about the differences between existing and new pre-training tasks, tracing model predictions to data points \citep{ilyas_datamodels_2022}, security questions such as membership inference \citep{carlini_membership_2022} and data poisoning \citep{geiping_witches_2021}, and a wide range of empirical investigations into topics such as stability or generalization that arise during training \citep{nagarajan_generalization_2019,jiang_fantastic_2019}. At the same time, we can imagine situations in which legal requirements make it unclear whether models trained on public data with uncertain origin are permissible, and where a practitioner is interested in retraining their language models using a specialized or trustworthy data source \citep{wilka_how_2017,gold_robots_2017}. 

In addition, we are motivated to benchmark the overall {\em conceptual} progress of research in this area over the last years, beyond simply turning the scaling knob. 
The goal of achieving BERT-like performance with modest training resources would have seemed unthinkable in 2018, and yet with modern advances and transformer training techniques this may now be possible.
%

\looseness -1 To answer these questions, we consider a challenge we call ``Cramming'' -- learning a whole language model the day before the test. Our studies begin by investigating many facets of the training pipeline to see which modifications actually improve performance in the scaled-down scenario. We provide evidence that even in this constrained setting, performance closely follows scaling laws observed in large-compute settings. An unsurprising consequence of these laws is that scaling down is hard; while smaller model architectures enable speeding up gradient computations, overall rates of model improvement over time remain nearly constant.  Nonetheless, we can find changes to the training recipe that exploit scaling laws to yield improvements by improving the effective rate of gradient computations without compromising model size.  In the end, we are able to train models that achieve respectable performance -- often close to and sometimes exceeding BERT on GLUE tasks -- on a shoestring budget\footnote{We provide code to replicate all experiments at \url{https://github.com/JonasGeiping/cramming}.}.

\section{Tying our hands behind our back: A setup with limited compute}\label{sec:rules}
Before we start this investigation, we want to outline the extent of limitations we are interested in.  The rules for cramming are as follows:
\begin{itemize}
    \item A transformer-based language model of arbitrary size is trained with masked-language modeling, completely from scratch. 
    \item Existing pretrained models cannot be included in any part of the pipeline.
    \item Any raw text (excluding downstream data) can be included for training. This means that one can achieve speedups by making judicious choices about how and when to sample data, provided the sampling mechanism does not require a pre-trained model. 
    \item The downloading and pre-processing of raw data is exempted from the total compute budget. Pre-processing may include CPU-based tokenizer construction, tokenization, and filtering, but cannot include representation learning (e.g. pre-training a word embedding is not allowed, unless it is counted towards the final runtime).
    \item Training proceeds on a single GPU for 24 hours. 
    \item Downstream performance is evaluated on GLUE \citep{wang_glue_2018}. Downstream finetuning on GLUE is limited to brief training with only the training data of the downstream task (we consider 5 epochs or less) and needs to work with hyperparameters set globally for all GLUE tasks. Downstream finetuning is excluded from the total compute budget.
\end{itemize}
In our implementation, we analyze both a setup with a classical \texttt{rtx2080ti} GPU (released September 2018) and separate setups with a more modern \texttt{rtxa4000} or \texttt{rtxa6000} GPU (released October 2020). We pair each unit with 4 CPU cores and 32GB of RAM.

Why these limitations? We are principally interested in re-investigating the original BERT setup of \citet{devlin_bert_2019} with limited compute. The optimal architecture of the transformer is not fixed, as the optimal size and shape depends on scaling laws \citep{kaplan_scaling_2020}. The limitations on usage of existing models rule out distillation from an existing model \citep{turc_well-read_2019,jiao_tinybert_2020,sun_mobilebert_2020,wang_minilm_2020,kaliamoorthi_distilling_2021} and data filtering based on existing large models \citep{golchin_compact_2022}, both of which ultimately answer questions about compression and transfer of already processed information. Further, we do not want to limit data to the original dataset used to train BERT, wanting to allow for possible improvements through better data curation and quality. The \texttt{rtx2080ti} GPU is a natural candidate for this experiment, given that it was released before \citet{devlin_bert_2019}, but the more recent \texttt{rtxa4000} is also interesting, as a more recent consumer-grade workstation variant. Finally we also test the \texttt{rtxa6000}, being arguably the upper limit of a single-user workstation. At the finetuning stage we want to mimic the original BERT finetuning and evaluation setup, but provide additional limits to prevent gains based on tuning of only the downstream procedure, for example via computationally extensive downstream training \citep{bahri_sharpness-aware_2021}, use of multiple downstream datasets (for example continued pretraining with MNLI before finetuning other tasks \citep{izsak_how_2021}), and extended hyperparameter optimization for each GLUE task \citep{devlin_bert_2019,liu_roberta_2019,lan_albert_2019}.

\section{Related Work on Efficient Transformers}

\begin{table}
    \centering
    \begin{tabular}{c|c|c|c|c}
    Group  & Target & Accelerator & Time Limit & Total exaFLOP \\ 
    \hline
    \citep{devlin_bert_2019}  & BERT & $16$ \texttt{TPU} & 4 days & 680 \\ 
    \citep{dettmers_tpus_2018} & BERT & $8$ \texttt{V100} & 11 days & 950 \\ 
    \citep{narasimhan_nvidia_2019} & BERT-large & 1472 \texttt{V100} & 47 min & 519 \\ 
    \citep{raffel_exploring_2020} & T5-base & $16$ \texttt{TPUv3} & 1 day & 170 \\ 
     \citep{iandola_squeezebert_2020} & squeezeBERT &  8 \texttt{Titan RTX} & 4 days & 361 \\ 
    \citep{narang_transformer_2021} & T5 variations & $16$ \texttt{TPUv3} & 1.75 days & 298
    \\ 
    \citep{tay_scale_2021} & T5-small-L16  & $16$ \texttt{TPUv3} & 11.2 hours & 82
    \\ 
   
    \citep{izsak_how_2021} & BERT variation & $8$ \texttt{V100} & 1 day & 86
 \\ 
    \hline 
    \citep{liu_roberta_2019} & roBERTa-base  & $1024$ \texttt{V100} & 1.25 day & 13\,824 \\
    \citep{chowdhery_palm_2022} & PaLM & $6144$ \texttt{TPUv4} & 50 days & 7\,299\,072 \\ 
    \hline
    Our Setup 1 & BERT variation & $1$ \texttt{rtx2080ti} & 1 day & 5 \\ 
    Our Setup 2 & BERT variation & $1$ \texttt{rtxa4000} & 1 day & 8 \\ 
    Our Setup 3 & BERT variation & $1$ \texttt{rtxa6000} & 1 day & 13 \\ 
    \end{tabular}
   
    \caption{Maximal Throughput available for select training runs of large language models. FLOP Counts for BERT reproductions and related models. Large-scale LMs included only for reference.}
    %
    %
    %

    \label{tab:flops}
\end{table}



%
%
%
%

\paragraph{How long does it take to train BERT?} 
\looseness -1 In general, this question is hard to answer, due to wildly varying hardware and software setups and differing measures of efficiency \citep{dehghani_efficiency_2021}. An upper bound on the compute of a training run can be established by finding the total number of (low-precision) floating point operations available over the wallclock budget of the run. This peak of total FLOPs in a given time interval is generally not reached in actual compute, even for highly optimized models \citep{chowdhery_palm_2022}, but represents the paid budget required to realize a training run. We summarize budgets for a few select training runs in \cref{tab:flops}. After the original training run for BERT on TPUs, initial reactions estimated up to 11 days of compute for comparable results on GPUs \citep{dettmers_tpus_2018}. However, sustained improvements, especially in software, have reduced the upper limit significantly \citep{you_large_2019,narasimhan_nvidia_2019}. Yet, recipes and implementations generally require entire server nodes (for GPUs) or TPU slices and target larger BERT architectures.

Other work discussing improvements to BERT targets compute settings closer to the original BERT, for example SqueezeBERT \citep{iandola_squeezebert_2020} employs 8 \texttt{Titan RTX} cards for four days. \citet{sellam_multiberts_2022} note that the original BERT training run is an outlier and doubling its training time more reliably reproduces the original results.

Our central point of comparison for BERT training with limited resources is the work of \citet{izsak_how_2021} who also attempt the goal of training BERT within 24 hours with overall similar limitations, but use a full server node with 8 \texttt{V100} GPUs. \citet{izsak_how_2021} choose a $\text{BERT}_\text{LARGE}$ architecture variant and train with sequence length of $128$, including a range of tweaks such as modified learning rates schedules, large batch sizes, sparse prediction and packed sequences. We re-evaluate this setup as a baseline setting for our own compute budget (which is about 15x smaller).

\paragraph{Studies of Efficient Transformers}
\looseness -1 Recent years have seen a flurry of research working to improve and modify the transformer architecture proposed in \citet{vaswani_attention_2017} and we refer to \citet{treviso_efficient_2022} for a recent categorization and review of research in this area. Several meta-studies have investigated proposed improvements and modifications: \citet{narang_transformer_2021} evaluate a large range of architectural modifications applied to the T5 model pipeline of \citet{raffel_exploring_2020} on tasks in both language understanding and translation. The encoder-decoder structure of T5 is closer in spirit to the original transformer setup, but is understood to behave similarly to BERT when using the encoder component \citep{liu_enct5_2021}. Evaluating modifications with 1.75 days of compute on \texttt{TPU} slices they find that most improvements do not reliably materialize gains in final accuracy. 
\citet{tay_scale_2021} work in the same setting and evaluate the optimal shape of T5 derived architectures and its relative effects on downstream performance as models are scaled. Further exploration of the scaling behavior of various architectural improvements in \citet{tay_scaling_2022} find that only few modifications outperform the original architecture of \citet{vaswani_attention_2017} at all scales, especially when evaluating downstream accuracy. The meta-study investigating improvements in preparation for extreme-scale training in \citet{scao_what_2022} focuses on minor modifications to layout, positional embeddings and data sources for autoregressive models, and other extremely-large scale training runs have so far been similarly conservative in their settings \citep{brown_language_2020,black_gpt-neox-20b_2022,rae_scaling_2022}.

In general though, these evaluations target larger compute settings than we intend to use, and are concerned with whether improvements (often from academic sources and proposed with evaluations on small scales) translate to larger scales. In this work, we set aside the question of (up)scaling and focus only on the limited compute. 




%
%
%

\paragraph{Scaling Laws}
The difficulty in finding tangible improvements is echoed in the scaling laws of \citet{kaplan_scaling_2020}. Over a wide range of transformer model shapes, \citet{kaplan_scaling_2020} find only model size (as number of parameters in non-embedding layers) strongly predicts performance. Further, for a fixed compute budget, an optimal model size can be derived, but performance is only mildly connected to model size - larger models processes less data per unit of compute, but improve faster by almost the same margin. While the precise coefficients and shape of these scaling laws continue to be iterated on \citep{hoffmann_training_2022} and adapted for related settings \citep{bansal_data_2022,clark_unified_2022,bahri_explaining_2021}, their overall logic appears hard to escape, even if power laws fit observations somewhat less well on small scales.
%


%
%

\section{Investigations}
For our experimental evaluation we implement and test a considerable number of proposed modifications to the setup of \citet{devlin_bert_2019} for their merits in our limited compute setting as described in \cref{sec:rules}. We first clarify the common implementation and initial data setup, and then investigate architectural, training and dataset improvements.

\subsection{Implementation Details}
We implement everything in PyTorch \citep{paszke_automatic_2017} and to limit our gains from the "software lottery" \citep{hooker_hardware_2021} we do not use specialized implementations, which would further bias results towards well-established components. We keep everything on the implementation level of the PyTorch framework, allowing only automated operator fusion \citep{sarofeen_introducing_2022} that can be applied to all components. Only after choosing a final architecture variant, we then re-enable the efficient attention kernel described in \citet{dao_flashattention_2022}. We run all experiments and ablation studies with the same setup of automated mixed precision \citep{micikevicius_mixed_2018} for standard 16- and 32-bit floating point precision {\color{gray}(over full 32-bit float, scaled 16-bit \bitep{rasley_deepspeed_2020} and pure \texttt{bfloat16} \bitep{wang_bfloat16_2019}. We find no benefit from offloading \bitep{ren_zero-offload_2021,rasley_deepspeed_2020} in our setting.)}. 

%

\paragraph{Initial Data Setup}\label{sec:data1}
We start our investigation with a close analogue to the original raw text sources of \citet{devlin_bert_2019}, using a recent dump of the English Wikipedia (\texttt{20220301.en}) and English \texttt{bookcorpus}, noting the commentary of \citet{tan_what_2019,bandy_addressing_2021}. We force all text into lower-case, strip accents and non-ascii characters and create an English tokenizer from scratch based only on this data. We choose WordPiece with a vocabulary size of $2^{15}=32768$ \citep{wu_googles_2016}. {\color{gray} We found no significant change in performance with BPE \bitep{sennrich_neural_2016} or SentencePiece with Unigrams \bitep{kudo_subword_2018,kudo_sentencepiece_2019}. Smaller vocabulary sizes ($2^{12}, 2^{13}, 2^{14}$) resulted in worse performance, while larger vocabulary sizes ($2^{16}$) we not reliably better}. We pack tokenized data into randomized sequences of length $128$ and separate unrelated fragments by \texttt{<sep>} {\color{gray} The performance impact from dropping this separator was minimal. No impact was observed from including a \texttt{<cls>} token in pretraining}. The shorter sequence length is sufficient for the downstream applications that we are targeting and simplifies attention computations. Packing data into full sequences limits us to simpler sequence losses, but uses the available compute optimally \citet{liu_roberta_2019,izsak_how_2021}.
For the targeted compute settings, this sequence length results in micro-batch sizes of $64$ to $96$ for most variations of the base BERT architecture on 
 the \texttt{gtx2080ti}, which we will accumulate into larger batch sizes. With our limited compute budget, this produces enough samples to run single-epoch training \citep{komatsuzaki_one_2019,hernandez_scaling_2022} where no data point is revisited.

\subsection{Modifying the Architecture}\label{sec:arch}
The most obvious way to efficiently scale down training is by modifying the model architecture; intuitively, it seems likely that smaller/lower capacity models will be optimal in the cramming regime. In this section, we study the relationship between model type and training efficiency.  We see that scaling laws create a strong barrier to scaling down.  Per-token efficiency of training depends strongly on model size, but not transformer type. Furthermore, smaller models learn less efficiently, and this largely mitigates any throughput gains. Fortunately, the fact that training efficiency is nearly constant across models of the same size means that we can boost performance by finding architecture modifications that speed up gradient computation while keeping the parameter count nearly constant.  This makes architecture selection fairly straightforward as we can make design choices based primarily on how they affect computation time for a single gradient step.

\paragraph{Scaling laws hold in the low-resource regime}
\looseness -1 A large corpus of research in recent years has developed architectural improvements to speed up the original transformer. 
Many of these methods have not been found to improve training for the large-scale T5 architecture \citet{narang_transformer_2021,tay_scaling_2022}.
But, in the low compute setting where data throughput is of utmost importance, maybe this is the way forward?  
Scaling laws have been observed by \citet{kaplan_scaling_2020} in the high-resource regime, and seem to hold strongly in the limit as resources grow. Surprisingly, these laws also hold in the limit of extreme compute down-scaling, and they create a barrier to low-cost training.

\begin{figure}
    \centering
    \includegraphics[width=0.49\textwidth]{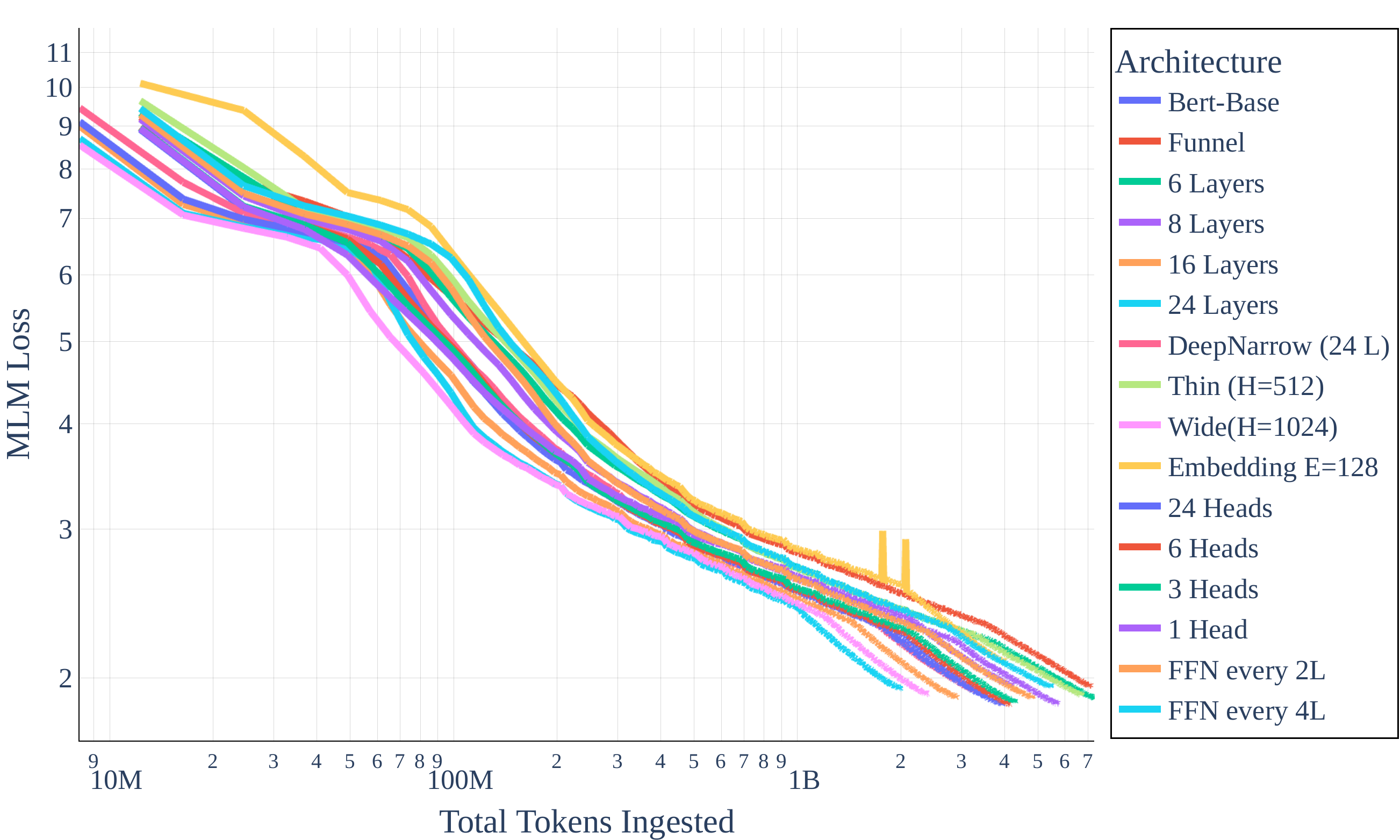}
    \includegraphics[width=0.49\textwidth]{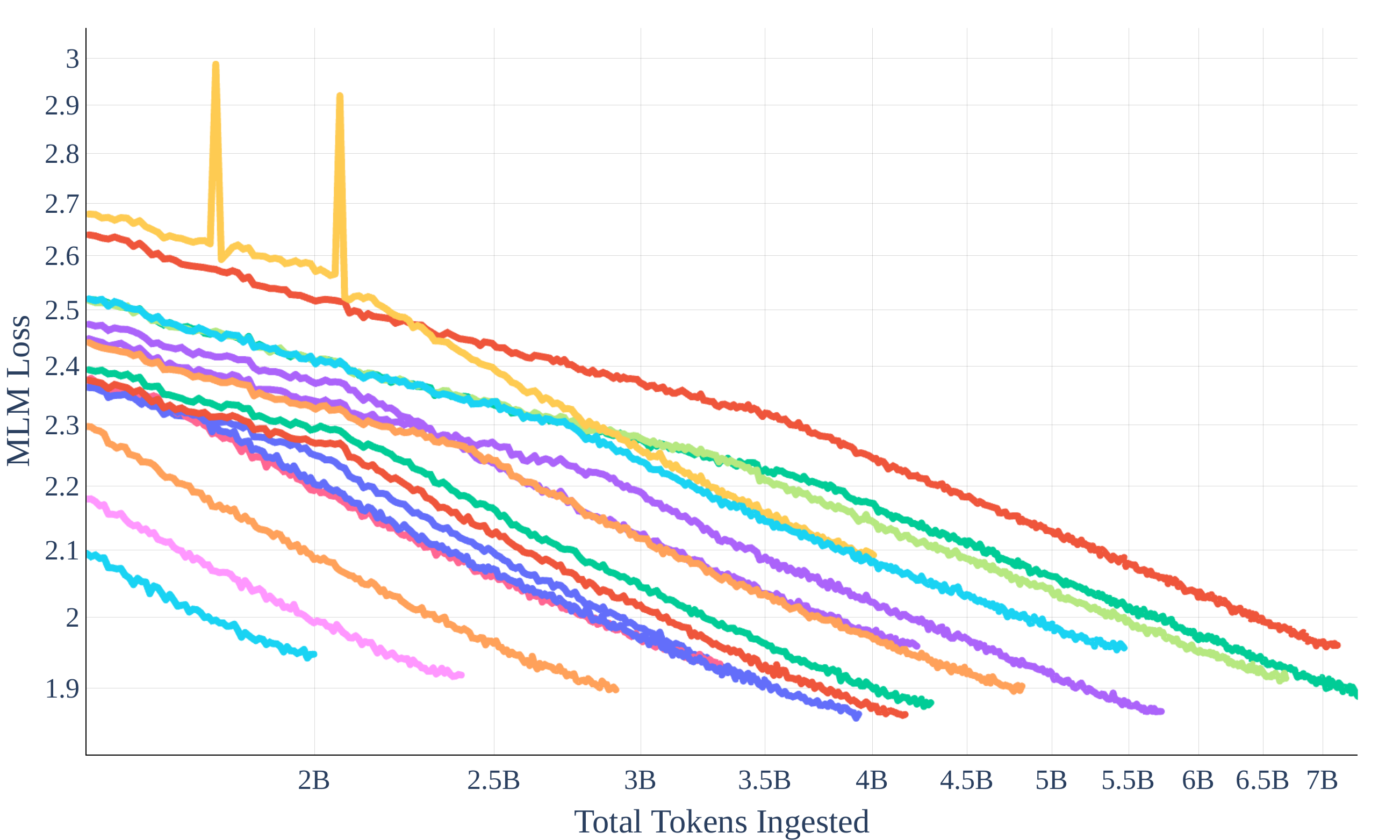}
    \caption{Various Transformer architectures and shapes, showing MLM loss versus number of tokens ingested. Left: Global view. Right: Zoom onto 10e8 or more tokens. All models trained with the same budget. We see that improvements through architectural reshaping are minimal; while there are some fluctuations in loss early in training, the rates of loss decay during most of training differ by a multiplicative constant (horizontal shift due to logarithmic horizontal axis) that depends strongly on the model size and not model type. }
    \label{fig:architecture}
\end{figure}

We exemplify the effect of scaling laws for many transformer variants from the literature in \cref{fig:architecture}, where we train each architecture variant with optimized training hyperparameters as described below in Section \ref{sec:training}. We apply these architecture variants to a shared baseline model that incorporates Pre-Normalization and rotary embedding. \cref{fig:architecture} visualizes the progress of MLM loss versus the number of tokens ingested in total and all architectures run with the same time budget. 

We observe that varying the transformer type and size has only minimal impact on the final loss after 24 hours. Models with more parameters learn more efficiently, as their MLM loss decreases faster on a per-gradient basis. However, smaller architectures make up for their slower learning efficiency by higher throughput, and thus process more tokens over the limited budget. \cref{fig:architecture} shows that different architectures are unpredictable throughout an initial stage of training (the first 1B tokens), after which the per-token efficiencies differ by only a multiplicative constant (a horizontal shift due to the log axis).  This constant depends almost entirely on the model size, not model type, so that all choices reach a MLM loss around $1.9$ at the end of training.

\paragraph{Exploiting the scaling law.}
The scaling laws seem to bar us from making large gains via major changes to the transformer size and type, as per-token performance is tightly coupled to model size.   ~{\color{gray} As a result,  we find no improvements when using a funnel-transformer architecture \bitep{dai_funnel-transformer_2020,nawrot_hierarchical_2022}, when dropping FFN layers \bitep{sridhar_trimbert_2022}, or when using recurrent layers \bitep{lan_albert_2019}, even when trained with BPTT as in \bitet{schwarzschild_easy--hard_2021}. Rescaling architectures to be deep-narrow \bitep{tay_scale_2021,wies_which_2021} provides no gains}.


While this principle closes one door for scaling down efficiently, it opens another;  Because per-gradient efficiency remains nearly constant for all models of the same size, we can exploit scaling laws by quickly searching for architectural choices that speed up computation while keeping model size roughly constant.  
A number of obvious optimizations fall into this category, and we describe them below, in addition to several other tweaks that provide marginal but worthwhile/free gains.


\paragraph{Attention Block:}
We disable all QKV biases \citep{dayma_dalle_2021}.  This exploits the scaling law by removing a layer of computation, making the forward and backward pass somewhat faster, while keeping the model size nearly constant.
We find that we could decrease gradient costs by reducing the number of attention heads \citep{merity_single_2019,araabi_optimizing_2020,liu_multi-head_2021,javaheripi_litetransformersearch_2022}, as this parallelizes better on the GPU and provides a slight performance boost. Yet, reducing the amount of heads also decreases finetuning performance, so we ultimately keep all $12$ heads. {\color{gray} We find no benefits from replacements to the softmax operation \bitep{richter_normalized_2020}.} 
We further keep the original multi-head self-attention mechanism. {\color{gray} A large amount of work has been focused on efficient attention \bitep{sukhbaatar_adaptive_2019,beltagy_longformer_2020,wang_linformer_2020,liu_tuformer_2021} and studies of efficient attention \bitep{tay_long_2020,tay_efficient_2020}. But, because we set the maximal sequence length to $128$, attention complexity is less of a concern in our setting. To verify this, we implement the recently proposed FLASH mechanism \bitep{hua_transformer_2022}, but find no benefits. We further experiment with Fourier attention as proposed in \bitet{lee-thorp_fnet_2021}, but find no improvements.
We find rotary embeddings \bitep{su_roformer_2021,black_gpt-neox-20b_2022}, to provide small benefits, but these are evened out by the drop in speed, so we ultimately decide against these.
}

\paragraph{Feedforward Block:} 
We find empirical gains from disabling all linear layer biases \citep{dayma_dalle_2021}.  Just as for the attention layers, this leverages the scaling law by accelerating gradient computation without noticeable impacts on model size.  As a result, we get higher throughput without compromising the rate at which the model improves.
We keep the original feedforward block largely unchanged, {\color{gray} finding no benefits from changing to another activation than GELU.} We do see small improvements from re-ordering the block into a gated linear unit \citep{dauphin_language_2017}. In contrast to other work, e.g. \citep{black_gpt-neox-20b_2022}, we do not increase the number of parameters in the FFN block to compensate for the halving of the hidden dimensionality due to gating.

\paragraph{Embedding:}
We implement scaled sinusoidal positional embeddings as described in \citet{hua_transformer_2022}, finding incremental benefits over learned or unscaled sinusoidal embeddings.
{\color{gray} We see no improvements from decoupling the input and output embeddings \bitep{chung_rethinking_2020}. The suggestion from \bitet{lan_albert_2019} to factorize the input embedding provides no gains in our setting.} We include a layer normalization at the end of the embedding block.

\paragraph{Layer Structure:}
\looseness -1 As observed in many studies, we find that pre-normalization with Layer Norms is beneficial over post Layer Norms \citep{baevski_adaptive_2018,xiong_layer_2020}. {\color{gray} We see no additional benefit from other variants of this modification, such as \bitep{liu_understanding_2020,shleifer_normformer_2021}. Further, replacing Layer Normalization with RMS Normalization provides no gains \bitep{zhang_root_2019}.} We note that the key effect of pre-normalization is to stabilize training and enable larger learning rates and reduced warmup, and we see limited benefits from including it by itself. {\color{gray} We see no benefits from stochastic dropping of entire layers as described in \bitep{zhang_accelerating_2020}}. 

\paragraph{Head Block:}
We find that we can remove the nonlinear head without ill effect. We can further drop the decoder bias \citep{radford_language_2019} and gain in memory using sparse token prediction \citep{liu_roberta_2019,izsak_how_2021}. We add a final Layer Norm to stabilize training further.





\subsection{Modifying the Training Setup}\label{sec:training}
We study the impact of training hyper-parameters on the BERT-base architecture.  The original BERT training recipe understandably results is poor model performance in the cramming setting, and so we revisit a number of standard choices.  

\paragraph{Objective:} We train with only masked language modeling on fully packed blocks of tokens with a masking rate of 15\% and the original setup of \citet{devlin_bert_2019} where $10\%$ of all masks are filled with random words and 10\% unchanged. {\color{gray} We see no improvement from masking at larger rates, e.g. at 40\% as proposed in \bitep{wettig_should_2022}, see Appendix. We see no difference enabling or disabling the mentioned 20\% rule. We evaluate other functions for the masked-language objective, such as mean-squared error \bitep{hui_evaluation_2021} or L1 loss, but find no benefits.}

\begin{figure}
    \centering
    \includegraphics[width=0.32\textwidth]{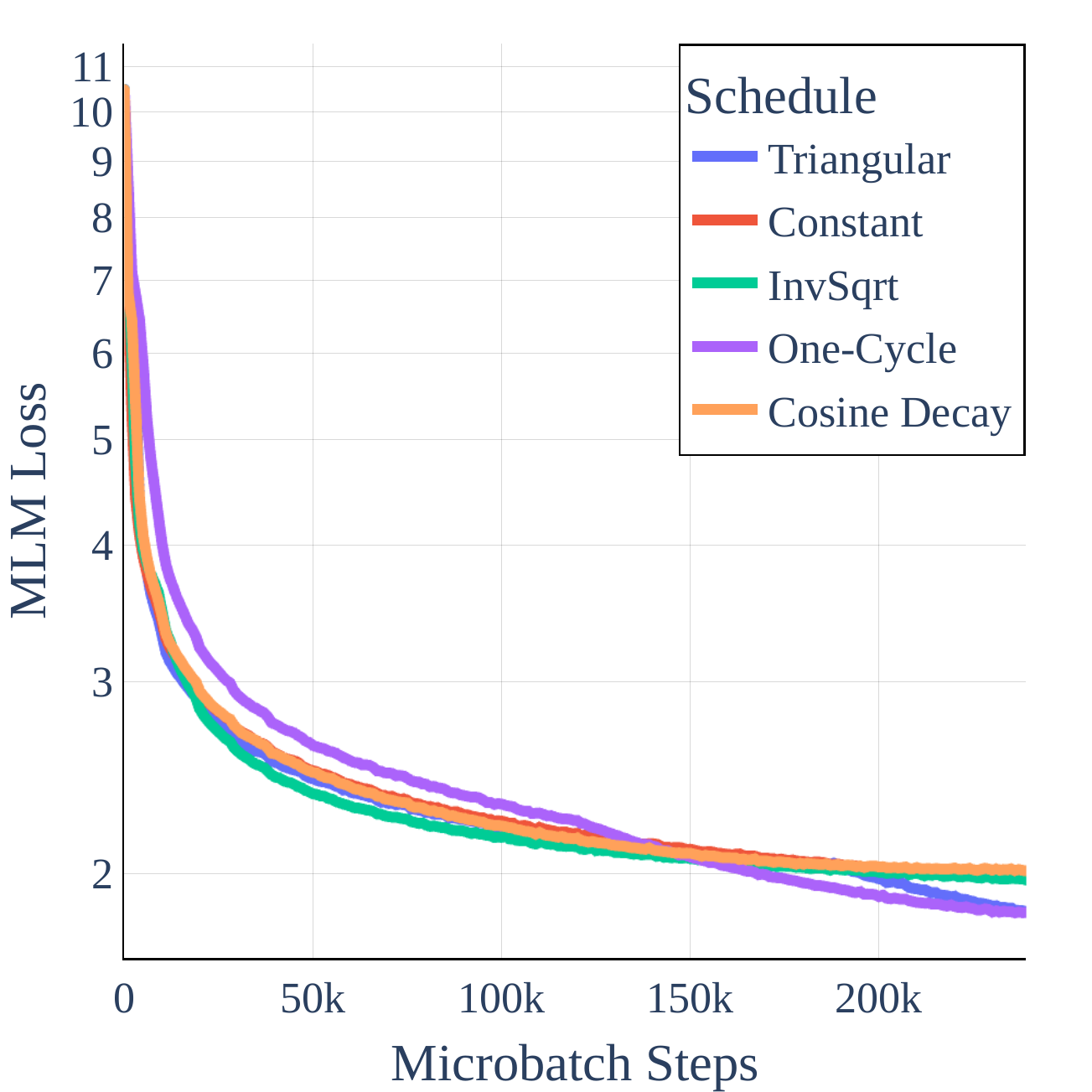}
    \includegraphics[width=0.32\textwidth]{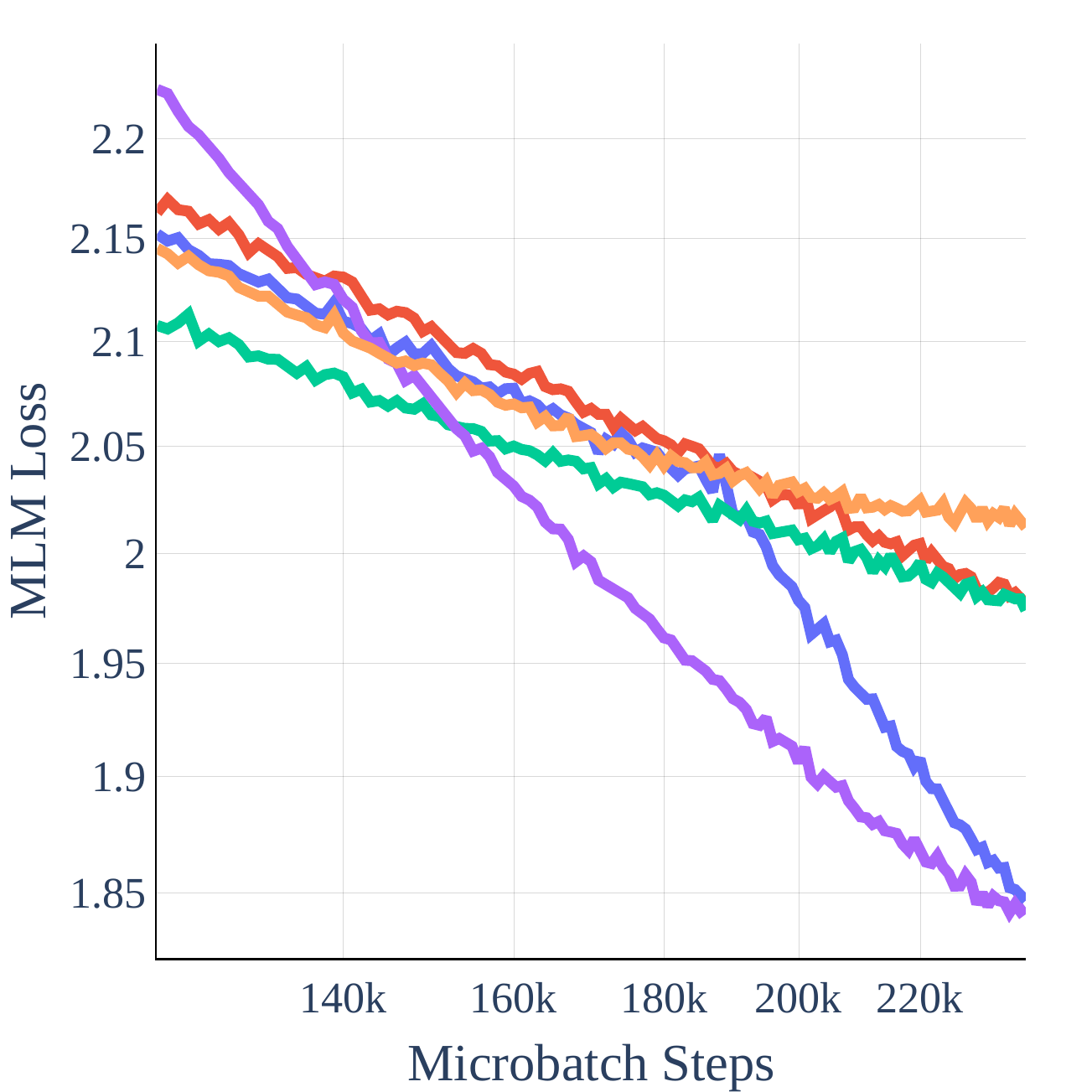}
    \includegraphics[width=0.32\textwidth]{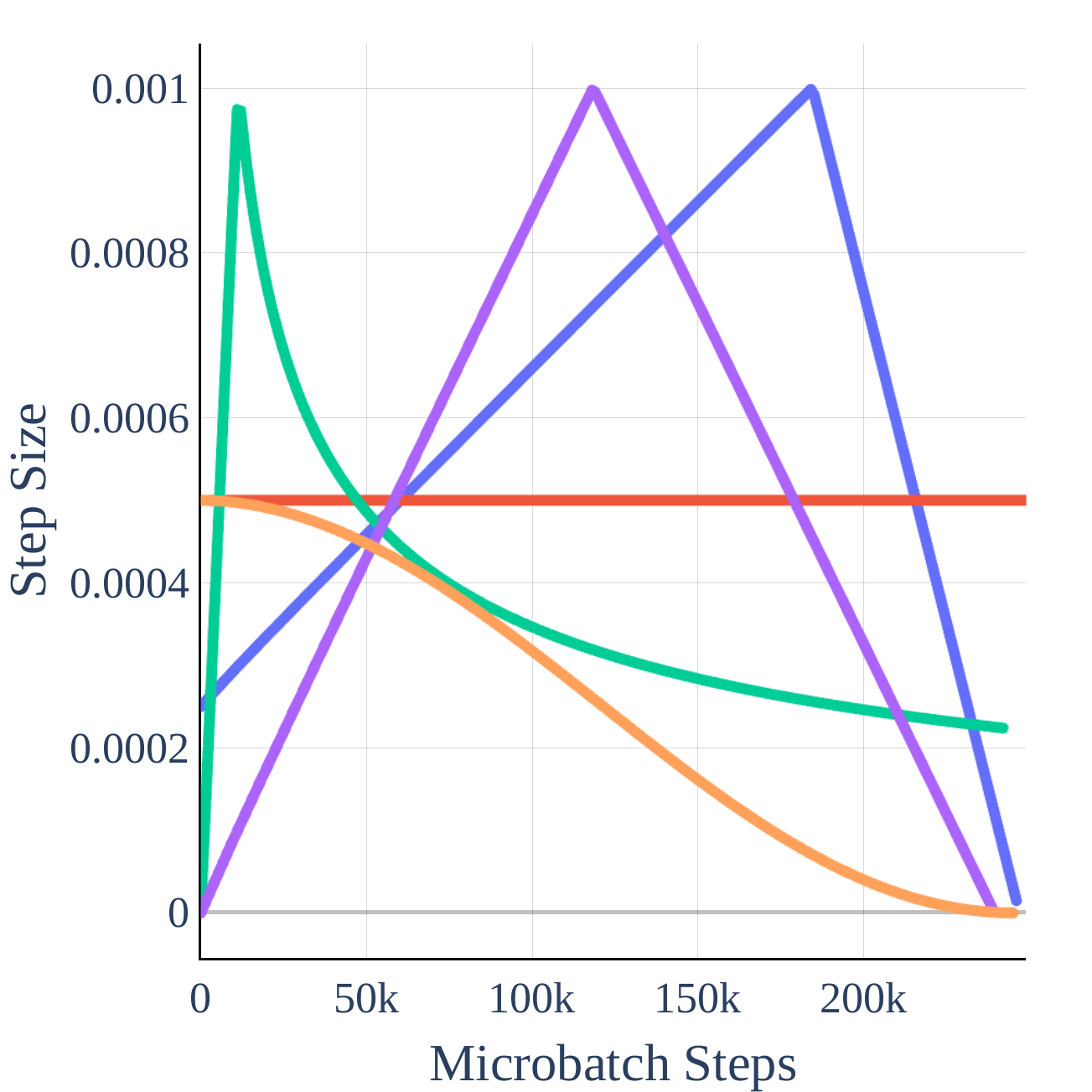}
    \caption{Learning Rate Schedules. Although globally many schedule result in similar behavior, we see in the zoom in the middle, that differences do exist. The right side shows the corresponding learning rate schedules. Both triangular-shaped one-cycle schedules have better end-time behavior, possibly due to the quick annealing.}
    \label{fig:learning_rates}
\end{figure}

\paragraph{Choice of Optimizer:}
We keep Adam \citep{kingma_adam:_2015} as the optimizer of choice, with weight decay of $0.01$ as described in \citep{loshchilov_decoupled_2017}, $\beta_1=0.9, \beta_2=0.98$ and $\varepsilon=\num{e-12}$. To stablize training at no extra cost, we include gradient clipping at a clip value of $0.5$. {\color{gray} We find no noticeable change in varying these parameters in reasonable amounts, e.g. $\varepsilon=\num{e-6}$, $\beta_1=0.9,\beta_2=0.999$. We test other first-order adaptive optimizers \bitep{shazeer_adafactor_2018, liu_variance_2020} but find no advantages in our setting. We further find no advantages using higher-order optimizers \bitep{yadav_making_2020,anil_scalable_2021}, but note that especially for higher-order optimizers there is a greater amount of variability in implementation.}

\paragraph{Learning Rate Schedule and Peak:}
Following the advice of \citet{izsak_how_2021}, we re-scale the learning rate schedule so that it is tied to our budget and the learning rate decays as the budget reduces to zero. 
Interestingly, we observe in \cref{fig:learning_rates} that while globally a large number of learning rate shapes lead to similar reductions in loss, we find that we can make some gains through the choice of schedule. We find that a simple one-cycle learning rate \citep{smith_super-convergence_2018} with a peak learning rate of $\num{e-3}$ leads to minimal pretraining loss within our budget.

\paragraph{Batch Size Schedule:}
\looseness -1 A particularity of our setting is that, due to being limited to a single GPU, the micro-batch size that finds its way onto this GPU ($96$ for most experiments) is several times smaller than the optimal batch size. We find that the optimal batch size in this setting is around $1536$ for minimal pretraining loss, but $4032$ for maximal downstream performance for the 2080ti, i.e. we accumulate gradients and only perform an update every 16 and 42 forward/backward passes, respectively. For the larger A4000 and A6000 cards, this corresponds to a micro-batch size of $128/256$ and final batch size of $4096$, which we again accumulate.

Fortunately, we can find small speedups by using an aggressive batch size schedule; we increase the number of averaged micro-batches linearly over the course of training. This results in more progress earlier in training, and leads to a small benefit to performance. 
{\color{gray} We also experiment with automatic and adaptive batching rules \bitep{de_big_2017,bollapragada_adaptive_2018,bollapragada_progressive_2018-1}, but find that the best results from these adaptive schedules resemble the fixed linear schedule.  For simplicity we just stick to the simpler linear schedule.}

\paragraph{Dropping Dropout}
The original BERT model of \citet{devlin_bert_2019} includes dropout as in \citet{vaswani_attention_2017}, which prevents overfitting when training data is small relative to total compute budget.  While it can be helpful as a regularizer, dropout effectively reduces the number of gradient updates seen by each parameter, as updates do not occur when the associated feature is dropped.  At the same time, update runtime is not strongly effected by the presence of dropout, and so dropout results in a net reduction in updates per second.

In the cramming setting, training data is large compared to compute. Overfitting is not possible due to the single epoch schedule, and we disable dropout during pretraining \citep{brown_language_2020} to maximize the number of parameter updates. We re-enable dropout during downstream fine-tuning with a dropout value of $0.1$.
{\color{gray} Further, we experiment with length curricula \bitep{li_curriculum_2022} (see appendix) and token dropping \bitep{hou_token_2022}, but find no gains in our setting. }


\subsection{Optimizing the Dataset}\label{sec:data}
\begin{table}
    \centering
\begin{tabular}{llr}
\toprule
Dataset & Batch Size & MNLI (m) \\
\midrule
Bookcorpus-Wikipedia & 1536 & 79.8 \\
The Pile & 1536 & 80.5 \\
The Pile (natural data subset) & 1536 & 80.8 \\
C4-Subset & 1536 & 79.1 \\
\hline
Bookcorpus-Wikipedia, Deduduplication $>100$ & 1536 & 79.9 \\
Bookcorpus-Wikipedia, Deduduplication $>50$ & 1536 & 79.5 \\
Bookcorpus-Wikipedia, filtered with $t=0.3$, sorted & 1536 & 80.8 \\
Bookcorpus-Wikipedia, sorted & 1536 & 81.0 \\ 
C4-Subset, Deduduplication $>100$  & 1536 & 79.2 \\
C4-Subset, filtered with $t=0.3$ & 1536 & 79.9 \\
C4-Subset, filtered with $t=0.3$, sorted & 1536 & 81.4 \\
C4-Subset, filtered with $t=0.3$, larger, sorted & 1536 & 81.9 \\ 
\hline
Bookcorpus-Wikipedia & 4032 & 80.5 \\
C4-Subset, filtered with $t=0.3$ & 4032 & 82.2 \\ 
C4-Subset, filtered with $t=0.3$, sorted & 4032 & 82.5 \\ 
\hline
C4-Subset, filtered with $t=0.3$ & 8064 & 80.9 \\
%
%
\bottomrule
\end{tabular}
    \caption{Dataset Variations for the optimal model from \cref{sec:arch} and optimal training routine from \cref{sec:training}, modifying final batch size in conjunction with dataset format.}
    \label{tab:datasets}
\end{table}
We found above that scaling laws create a barrier to making major gains (beyond computational efficiencies) with architectural modifications.  However, scaling laws do not preclude us from training on better data.  Once we have exhausted our ability to train on more tokens per second, we should seek to train on better tokens.

We consider two data based pathways to better down-scaling. First, we can filter, process, or sort the existing data in various ways. Second, we can swap our data source. To this end, we experiment with several subsets of \textit{The Pile} \citep{gao_pile_2020}, containing raw text from only \textit{Gutenberg}, \textit{Books3} and \textit{Wikipedia (en)}. From these Pile datasets we tokenize the first \num{4e6} entries to generate enough tokens for our single pass. 
Another popular source of data is C4, the colossal, cleaned version of Common Crawl \citep{raffel_exploring_2020}, from which we stream the first \num{20e6} entries. For each data source we regenerate its own WordPiece tokenizer as described in \cref{sec:data1}.

Of these four sources, we find the Pile to perform best in terms of downstream MNLI performance. However, it turns out we can further improve especially the C4 datset through additional processing. We first evaluate deduplication as described in \citet{lee_deduplicating_2022} via exact substring deduplication, but find this not to help in downstream performance in our case. We then test filtering for uncompressible data. We use the tokenizer itself to remove all training sequences from C4 set that cannot be compressed well; we simply set a threshold $t$, e.g. $t=0.3$, and drop all entries from the dataset where the number of tokens in the entry is larger than $t$ times the number of raw characters. This removes, for example, sequences consisting of hard-to-compress HTML or markdown code. Surprisingly, this results in a measurable improvement on C4, summarized in \cref{tab:datasets}.

We then see some further improvements from two directions. First, sorting all tokenized sequences by some metric, and second, increasing the final batch size. For filtering we sort all tokenized sequences by their average (unigram) token prevalence, so that likely sequences occur first. This has some positive effect, and can be strengthened slightly by drawing from a larger corpus, as the unlikely sequences never get reached. 
Finally, increasing the batch size to $4032/4096$ at the end of training (as mentioned in \cref{sec:training}) is disproportionally effective on C4, but less so on \texttt{bookcorpus-wikipedia}. We believe that both modifications ultimately reduce the likelihood of training being hindered by fluctuations in the data distribution.

\paragraph{Vocabulary Size}
\begin{wrapfigure}[16]{r}{0.5\textwidth}
\vspace{-0.41cm}
    \centering
    \includegraphics[width=0.33\textwidth]{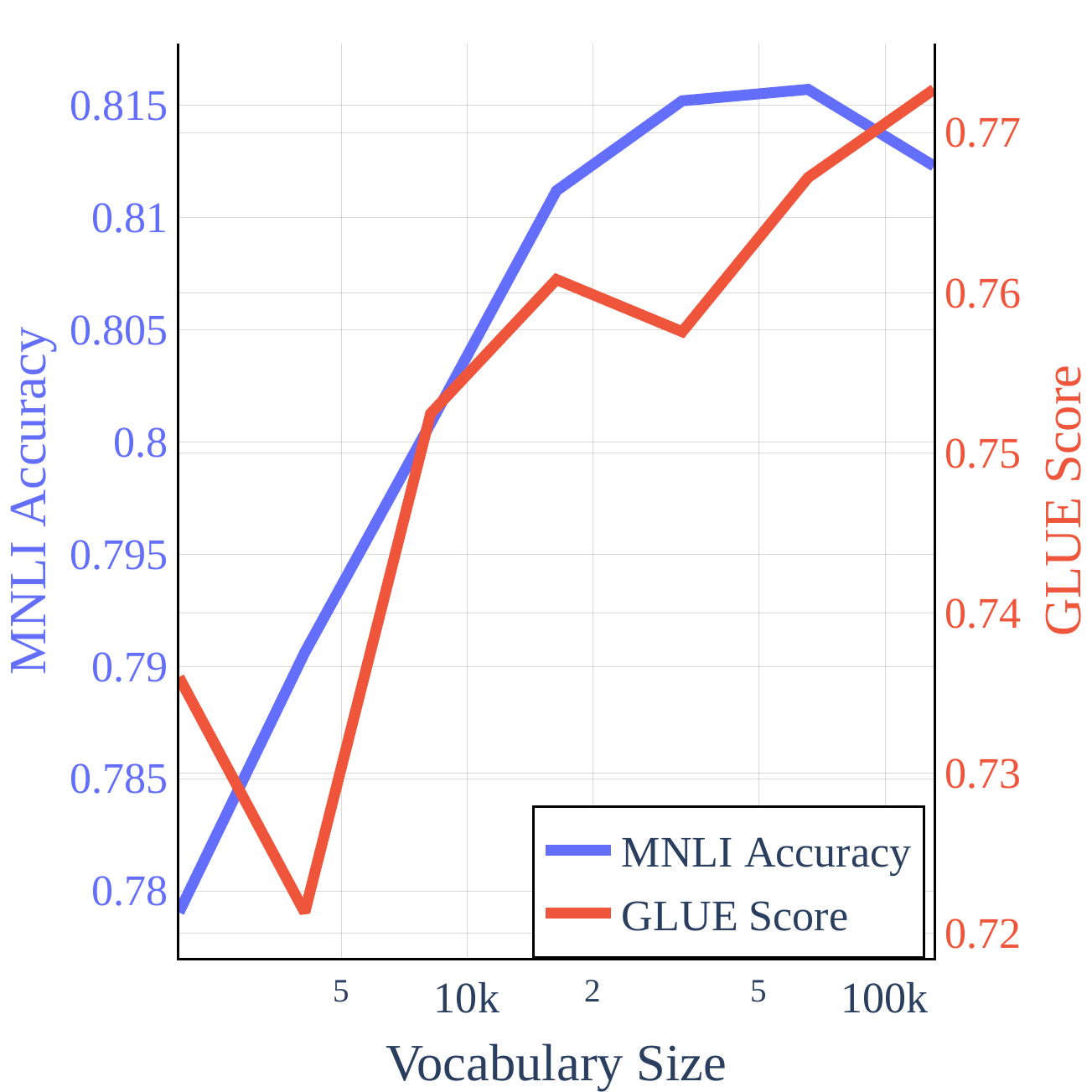}
    \caption{Vocabulary Size versus GLUE Score and MNLI Accuracy for models trained in the cramming regime on \texttt{bookcorpus-wikipedia} data.}
    \label{fig:vocab_size}
\end{wrapfigure}
We also check whether the original vocabulary size of $32768$ described in \citep{devlin_bert_2019} is optimal in the crammed regime. A priori, this might not hold: The larger, the vocabulary, the more, unique tokens and relationships between unique tokens have to be learned during training. On the other hand, increasing the vocabulary size would compress data further (albeit vanishingly after some point), which would allow for more information to be compressed into the fixed number of tokens that can be ingested during the crammed training run. In \cref{fig:vocab_size}, we find that for \texttt{bookcorpus-wikipedia} data, larger vocabulary sizes correlate with larger average GLUE score, although the effect is plateauing for the MNLI task around the original $32768$ vocabulary size. Moving forward, we hence keep this vocabulary size.


%

\section{Finetuning Performance on GLUE}\label{sec:downstream}

Finally, we systematically evaluate performance on the GLUE benchmark of \citet{wang_glue_2018}, minus WNLI as in \citet{devlin_bert_2019}. We note that we only use MNLI (m) during the previous sections and do not tune hyperparameters based on the full GLUE scores. We finetune both the pretrained BERT-base checkpoint and our models under the same constraints laid out in \cref{sec:rules}. For BERT-base, we finetune all datasets for $5$ epochs with a batch size of $32$ and learning rate of \num{2e-5}. For the crammed models, we find that this is not optimal and minor improvements can be gained from a batch size of $16$ and learning rate of \num{4e-5} with cosine decay (this setup does not improve the pretrained BERT checkpoint). 

\begin{table}[t]
    \centering
    \small
    \addtolength{\leftskip} {-1cm}
    \addtolength{\rightskip}{-1cm}
\begin{tabular}{lrrrrrrrr|r}
\toprule
 & MNLI & SST-2 & STSB & RTE & QNLI & QQP & MRPC & CoLA & GLUE \\
\midrule
BERT-base (Fully trained) & 83.2/83.4 & 91.9 & \textbf{86.7} & \textbf{59.2} & \textbf{90.6} & \textbf{87.7} & \textbf{89.3} & \textbf{56.5} & \textbf{80.9} \\
BERT-base (No Pretrain) & 34.1/34.1 & 79.9 & 17.8 & 47.3 & 50.0 & 68.6 & 77.9 & 0.0 & 45.5 \\
\hline
& \multicolumn{9}{c}{Trained for 1 day on a \textbf{2080ti}:}\\
\hline
BERT (normal protocol) & 58.7/57.8 & 79.8 & 16.6 & 50.9 & 55.4 & 71.1 & 70.1 & 7.3 & 52.0 \\
BERT (\citep{izsak_how_2021}) & 75.0/75.7 & - & - & 52.3 & 84.6 & 84.4 & 82.2 & 33.8 & 69.7 \\
crammed BERT & 82.8/83.4 & 91.5 & 83.1 & 54.0 & 89.0 & 87.2 & 86.2 & 47.2 & 78.3 \\
\hline
& \multicolumn{9}{c}{Trained for 1 day on an \textbf{A4000}:}\\
\hline
BERT (normal protocol) & 58.0/56.5 & 79.4 & 17.0 & 51.6 & 54.2 & 70.6 & 74.1 & 8.2 & 52.2 \\
BERT (\citep{izsak_how_2021}) & 58.8/59.6 & - & - & - & - & - & 81.4 & 0.0 & 49.9 \\
crammed BERT & 83.0/83.2 & 91.6 & 84.8 & 54.7 & 88.5 & 86.9 & 86.4 & 43.7 & 78.1 \\
\hline
& \multicolumn{9}{c}{Trained for 1 day on an \textbf{A6000}:}\\
\hline
BERT (normal protocol) & 56.3/54.8 & 81.2 & 21.8 & 49.5 & 56.4 & 65.1 & 74.8 & 10.3 & 52.2 \\
BERT (\citep{izsak_how_2021}) & 76.2/76.5 & 87.4 & 78.5 & 49.1 & 85.0 & 84.1 & 83.2 & 36.3 & 72.9 \\
crammed BERT & \textbf{83.9}/\textbf{84.1} & \textbf{92.2} & 84.6 & 53.8 & 89.5 & 87.3 & 87.5 & 44.5 & 78.6 \\
\bottomrule
\end{tabular}
    \caption{Comparison in GLUE-dev performance of baseline BERT to the crammed model. Note that all runs abide by the finetuning protocol described in \cref{sec:rules} with fixed hyperparameters for all tasks and an epoch limit of 5. Missing values are NaN. The protocol of \citep{izsak_how_2021} was designed for an 8 GPU server blade, and it crammed onto a single GPU here. The MNLI column shows evaluation results for both matched and mismatched sets. The GLUE column depicts the full average over the same tasks as in \citet{devlin_bert_2019}.
    }
    \label{tab:avg}
\end{table}

\Cref{tab:avg} and \cref{tab:glue} describe the performance of this setup on the GLUE downstream tasks (as median over 5 downstream trials). There we compare the original BERT-base checkpoint, a reproduction of the BERT pretraining settings stopped after our budget is reached, the setup described in \citep{izsak_how_2021} and the modified recipe, trained for a single day for each GPU setup. Overall, performance is surprisingly decent, especially for the larger datasets of MNLI, QQP, QNLI and SST-2, where downstream finetuning can smooth out the remaining differences between the full BERT model and the crammed variants. Further, we find substantial gains over both a naive BERT training with limited budget, and over the recipe described in \citep{izsak_how_2021}. For \citep{izsak_how_2021}, the described recipe was originally designed for a full 8 GPU server blade, and squeezing the BERT-large model therein onto the smaller GPUs in this experiment is resposnsible for most of the performance degradation of this recipe in our scenario.

Overall, the crammed model mostly works, even for smaller datasets. The average is brought down however by a significant drop on CoLA (corpus of linguistic acceptability) \citep{warstadt_neural_2019}. 
\begin{wraptable}[12]{r}{0.5\textwidth}
\centering
\caption{Comparison in GLUE-dev performance of baseline BERT to crammed model. Avg. Score is all scores excluding CoLA, GLUE is the full average over the same tasks as in \citet{devlin_bert_2019}.}
\label{tab:glue}
\centering
\small
\begin{tabular}{lrrr}
\toprule
 & CoLA & Avg. Score & GLUE \\
\midrule
Bert-Base & \textbf{56.5} & \textbf{84.0} & \textbf{80.9} \\
\hline
Crammed (2080ti) & 47.2 & 82.1 & 78.3 \\
Crammed (A4000) & 43.7 & 82.4 & 78.1 \\
Crammed (A6000) & 44.5 & 82.9 & 78.6 \\
\bottomrule
\end{tabular}
\end{wraptable}
This behavior is intriguing and we offer two hypotheses. First, it is conceivable that the chosen global hyperparameters for finetuning are a bad fit for CoLA in particular. CoLa performance can be brittle with respect to hyperparameter, with \citet{jiao_tinybert_2020} training longer only on CoLA or \citet{joshi_spanbert_2020} training less only on CoLA. Nevertheless, for BERT, a set of global hyperparameters exists, pointing at a deficiency in the crammed model. As a second hypothesis, it is conceivable that these models need to process more text before they memorize enough data to do well on CoLA. This would be in contrast to  \citet{liu_probing_2021}, who find that CoLA is learned relatively quickly compared to other downstream tasks when probing intermediate BERT checkpoints. On the other hand, deficiencies on CoLA in particular are also common in approaches that distill BERT into smaller architectures \citep{sun_patient_2019,turc_well-read_2019,mukherjee_xtremedistiltransformers_2021}, which might come with limited capacity for linguistic acceptability.



\subsection{Ablation - Which Changes Really Mattered?}
In \cref{tab:ablation_data_train_arch} we provide a summary ablation study of all changes discussed in this work. We group modifications, as in previous sections into the three groups of architecture, training and data and ablate each group by resetting all modifications to the original BERT recipe. Here, we find that we first have to make minimal modifications in any case, as modifications to architecture, such as PreNorm layer structures also in turn allow the more aggressive learning rate schedules described in the training setup. Taking this into account, we ultimately find about two percentage points gained in average GLUE score through architectural changes, one percentage point in data changes, and half a precentage point in training modifications.

\begin{table}
    \centering
    \small
\begin{tabular}{lrrrrrrrr|r}
\toprule
 & MNLI & SST-2 & STSB & RTE & QNLI & QQP & MRPC & CoLA & GLUE \\
\midrule
crammed BERT  & \textbf{83.9} / \textbf{84.1} & \textbf{92.2} & 84.6 & 53.8 & \textbf{89.5} & \textbf{87.3} & 87.5 & \textbf{44.5} & \textbf{78.6} \\
+ original data & 82.2 / 82.7 & 92.0 & 83.6 & 49.8 & \textbf{89.5} & 87.0 & 85.9 & 42.5 & 77.3 \\
+ original train & 50.0 / 50.4 & 80.7 & 13.7 & 52.0 & 59.8 & 65.1 & 73.2 & 7.2 & 50.2 \\
+ original arch. & 35.4 / 35.2 & 49.1 & - & 52.7 & 49.5 & 0.0 & 0.0 & 0.0 & 27.7 \\
\hline
+ minimal train mod. & 81.9 / 82.6 & 91.4 & \textbf{85.5} & \textbf{54.9} & 88.2 & 87.0 & \textbf{88.4} & 43.6 & 78.1 \\
+ minimal arch. mod. & 83.2 / 83.5 & 91.7 & 82.0 & 52.0 & 88.9 & 86.8 & 83.6 & 38.3 & 76.7 \\
\bottomrule
\end{tabular}
    \caption{Ablation study, which improvements were most important? The first group shows an ablation where one component of the final combination of training, architecture, and data modifications (the crammed BERT model) is replaced by the original setup. Here, we find that modifications in training and architecture have to co-occur. For example, the aggressive learning rate schedule can only be used when the model also contains pre-normalization Layer Norms. As such we also include a row with \textit{minimal training} modifications (dropout disabled, cosine decay to zero within budget with warmup, fixed batch size of 4096) and a row with \textit{minimal architecture} modifications (Pre-normalization, sparse activations, Layer Norm $\varepsilon=\num{e-6}$).
    }
    \label{tab:ablation_data_train_arch}
\end{table}

\subsection{What Happens When Training Longer?}
We also verify what happens if the cramming recipe discussed so far is used with more budget. To this end, we train models for 48 hours on 8 A6000 GPUs, which ends up to be $208$ total exaFLOP, c.f. \cref{tab:flops}. We directly apply the setting described so far, simply scaling the learning rate schedules to cover the new budget of 48 hours. In \cref{tab:more_compute}, we find that the discussed recipe does immediately generalize to larger compute budgets. This is surprising, not the least, as now, the dataset (which was sorted in \cref{sec:data} is now too small and repeated multiple times. The newly trained models have strong performances, especially on MNLI and SST-2, where they significantly outperform the original BERT checkpoint and fall into a similar range as the roBERTA-base checkpoint of \citet{liu_roberta_2019}, which was trained with much more compute. Yet, in other tasks, such as (again) CoLA, the new models barely improve even in the larger compute regime.

\begin{table}[]
    \centering
    \small
    \addtolength{\leftskip} {-1cm}
    \addtolength{\rightskip}{-1cm}
\begin{tabular}{lrrrrrrrrrr}
\toprule
 & MNLI & SST-2 & STSB & RTE & QNLI & QQP & MRPC & CoLA & GLUE \\
\midrule
BERT-Base (Fully trained) & 83.2/83.4 & 91.9 & 86.7 & 59.2 & 90.6 & 87.7 & 89.3 & 56.5 & 80.9 \\
BERT-Base (No Pretrain) & 34.1/34.1 & 79.9 & 17.8 & 47.3 & 50.0 & 68.6 & 77.9 & 0.0 & 45.5 \\
ROBERTA-Base & \textbf{86.6}/86.4 & 93.7 & \textbf{90.4} & \textbf{77.3} & 92.1 & \textbf{88.3} & \textbf{91.4} & \textbf{60.2} & \textbf{85.1} \\
\hline
Crammed BERT (A6000) & 83.9/84.1 & 92.2 & 84.6 & 53.8 & 89.5 & 87.3 & 87.5 & 44.5 & 78.6 \\
\hline
& \multicolumn{9}{c}{Trained for 2 days on 8 \textbf{A6000}:}\\
\hline
Crammed BERT & 86.5/\textbf{86.7} & \textbf{93.8} & 86.8 & 53.4 & 91.6 & 88.0 & 88.2 & 42.9 & 79.8 \\
Crammed BERT (no clipping) & 86.1/\textbf{86.7} & 93.2 & 87.1 & 55.2 & \textbf{92.1} & 88.3 & 90.2 & 46.6 & 80.6 \\
\bottomrule
\end{tabular}
    \caption{Models trained on 16x as much compute as otherwise in this work, but with exactly the same setup and data. With this budget, we use about half as much compute as one of the original BERT training runs. The resulting models (which are surprisingly slightly improved by removing gradient clipping again), are equivalent in performance, even to ROBERTA-base models trained in \citet{liu_roberta_2019}, on some tasks. On other tasks, such as CoLA and RTE, the additional compute barely improves performance.}
    \label{tab:more_compute}
\end{table}

\section{Limitations}
In this work, we limited our investigation to transformer-based architectures trained with MLM objectives. However, we do think that the general task of cramming posed in \cref{sec:rules} is interesting even when relaxing these constraints. There have been a number of modifications proposed to the objective in particular \citep{joshi_spanbert_2020,bao_unilmv2_2020,bajaj_metro_2022,tay_unifying_2022}. While \citet{artetxe_role_2022} and \citet{wang_what_2022} find MLM still to hold up well as a pretraining objective, other suggestions such as ELECTRA  \citep{clark_electra_2019,clark_pre-training_2020,he_debertav3_2021} could be employed which might be beneficial for crammed models. Also, the optimal architecture might not be transformer-based \citep{merity_single_2019,fusco_pnlp-mixer_2022,peng_bo_rwkv-lm_2021}.

\section{Conclusions}
We discuss how much performance a transformer-based language model can achieve when crammed into a setting with very limited compute, finding that several strands of modification lead to decent downstream performance on GLUE.
Overall though, cramming language models appears hard, as we empirically find many implications of \citet{kaplan_scaling_2020} to still hold in this regime, and for examples improvements through larger models are evened out by their slower speed. 
We hope that this work can provide a baseline for explorations of the question of cramming we formalize in \cref{sec:rules} and cast an additional light on a number of improvements and tricks proposed for transformer architectures in recent years.





\subsubsection*{Reproducibility Statement}
We provide code to reproduce all experiments at \url{https://github.com/JonasGeiping/cramming}.



\bibliography{zotero_library}
\bibliographystyle{iclr2023_conference}

\appendix
\section*{\hrulefill Appendix \hrulefill}
%


\section{Other Modifications}
\looseness -1 A few recent developments not included in this study are \citet{roy_n-grammer_2022}, \citet{shen_staged_2022}, and \citet{mindermann_prioritized_2022}. Modifications further not included in this study are more involved initialization \citep{zhu_gradinit_2021}, additional objective modifications \citep{muller_when_2019}, progressive growth \citep{gu_transformer_2021,shen_staged_2022}, convolutional variants \citep{iandola_squeezebert_2020,chelombiev_groupbert_2021, so_searching_2021}, sequence recurrence \citep{lei_simple_2022} and TUPE embeddings  \citep{ke_rethinking_2020}.
%
%

\section{Additional Information}

Additional results concerning architecture modifications can be found in  \cref{tab:arch1} and \cref{tab:arch2}. Additional results for training modifications can be found in \cref{tab:train_raw}. Not all results remarked on in the main body (especially for variations that did not work, marked in gray) are accompanied by raw results in this appendix, but can be computed using the provided implementation.

\begin{figure}
    \centering
    \includegraphics[width=0.32\textwidth]{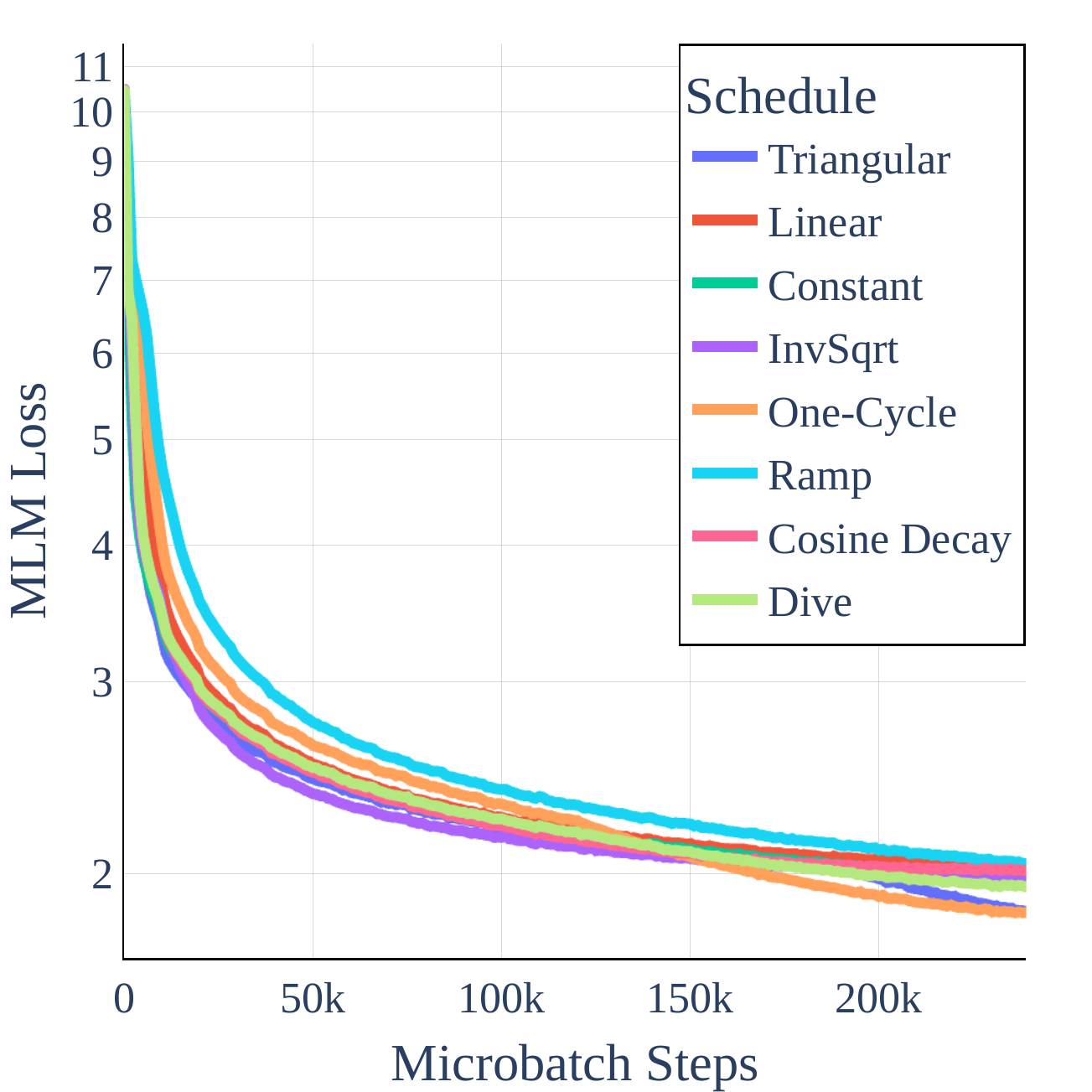}
    \includegraphics[width=0.32\textwidth]{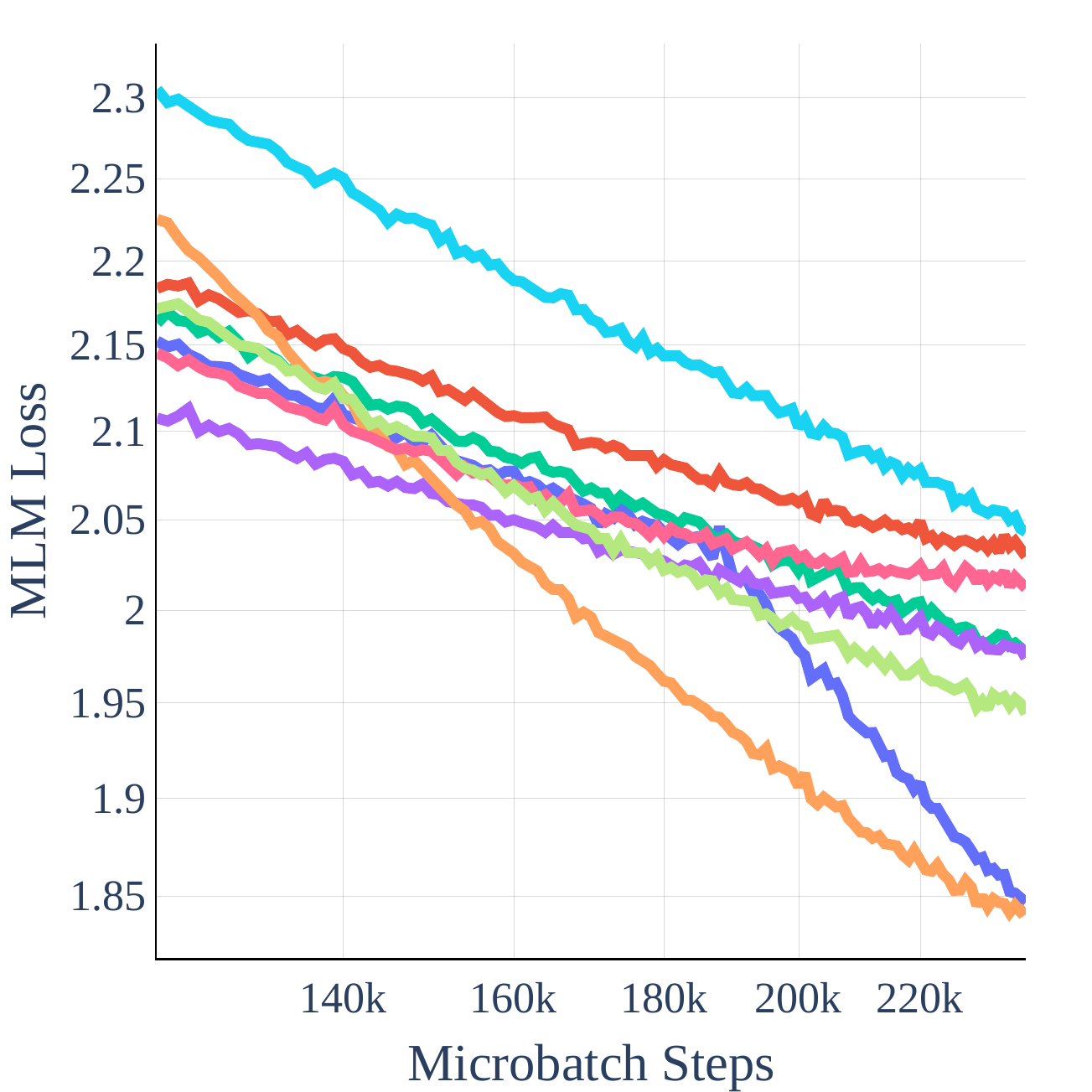}
    \includegraphics[width=0.32\textwidth]{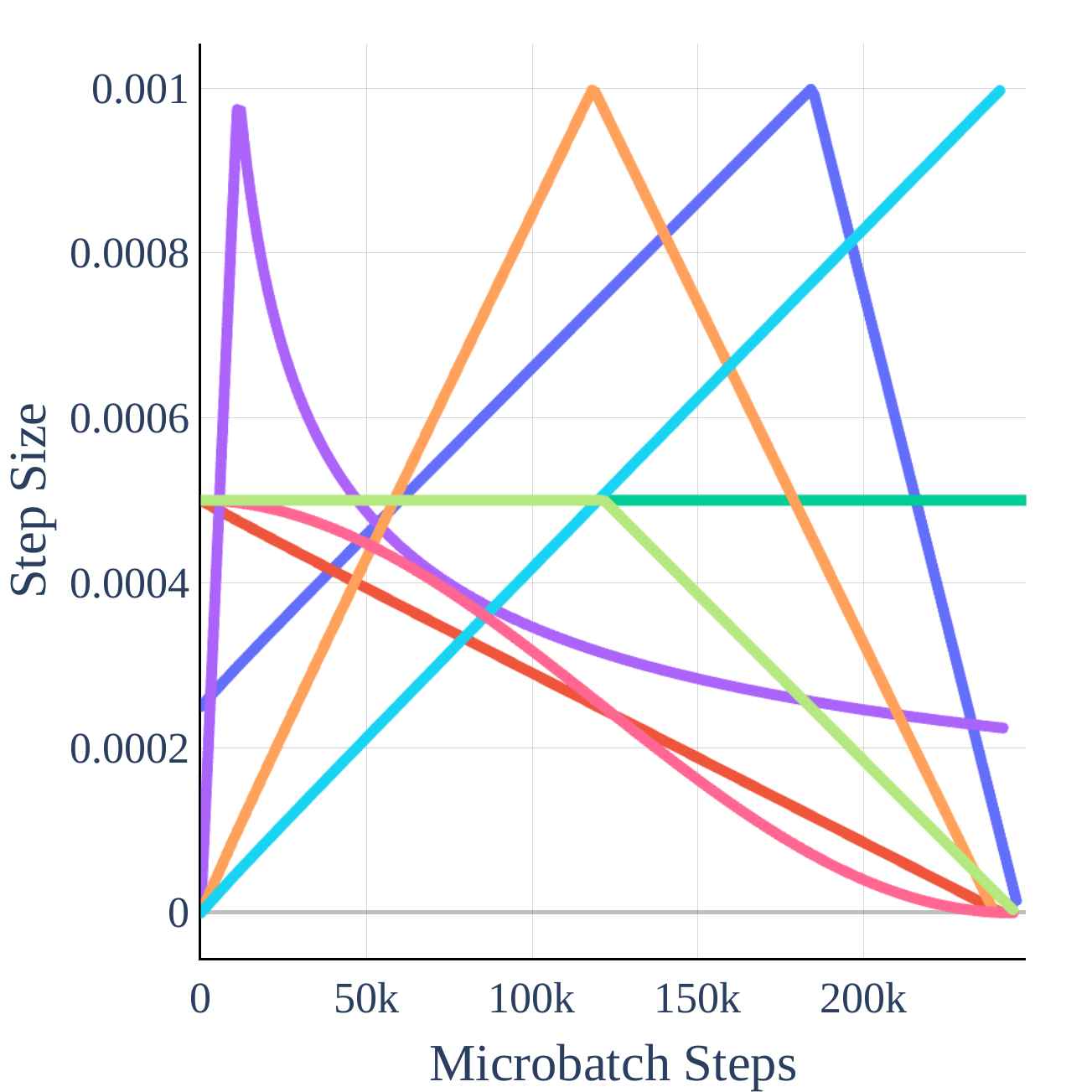}
    \caption{Extended version of \cref{fig:learning_rates}, including additional learning rate schedules.}
    \label{fig:learning_rates2}
\end{figure}

\begin{figure}
    \centering
    \includegraphics[width=0.49\textwidth]{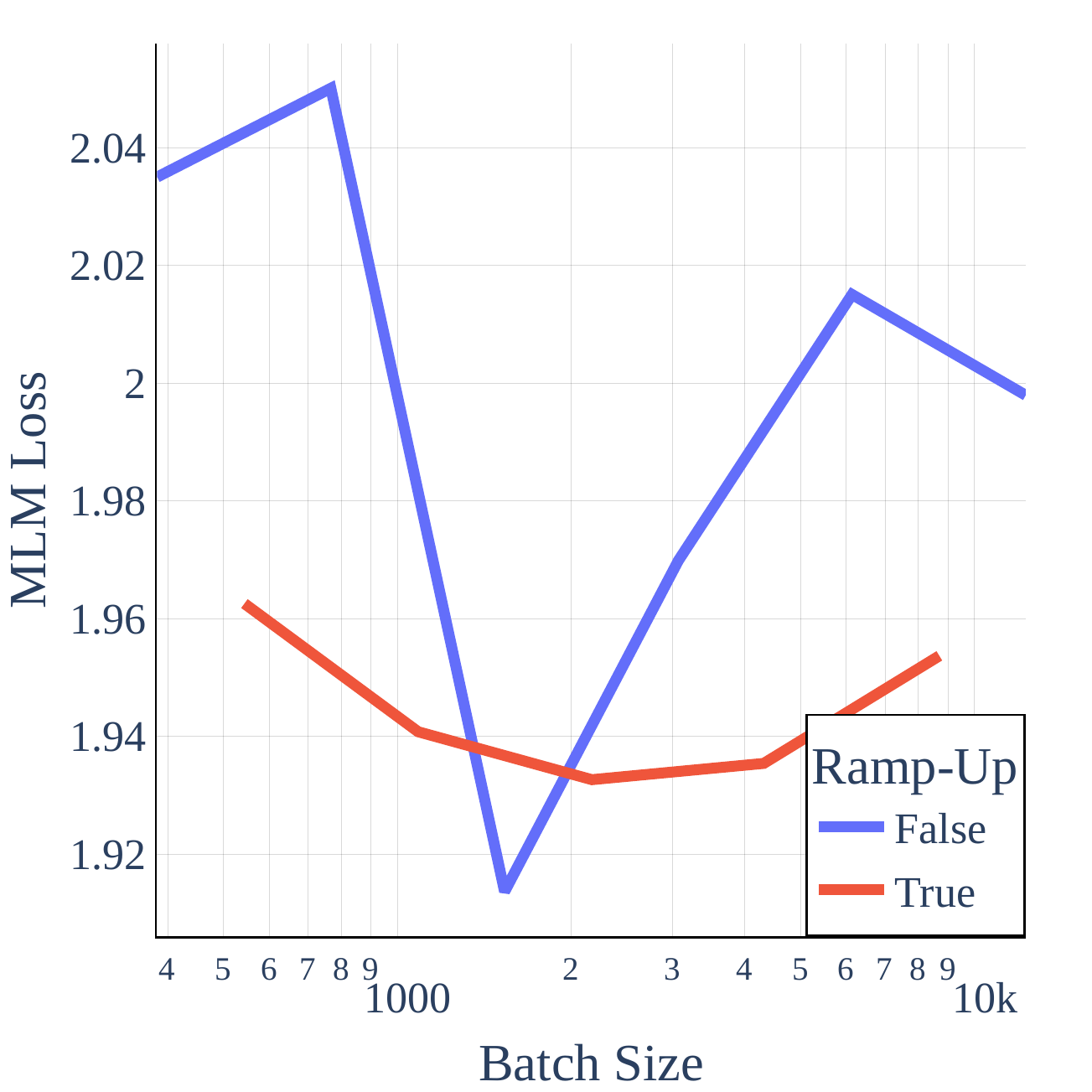}
    \includegraphics[width=0.49\textwidth]{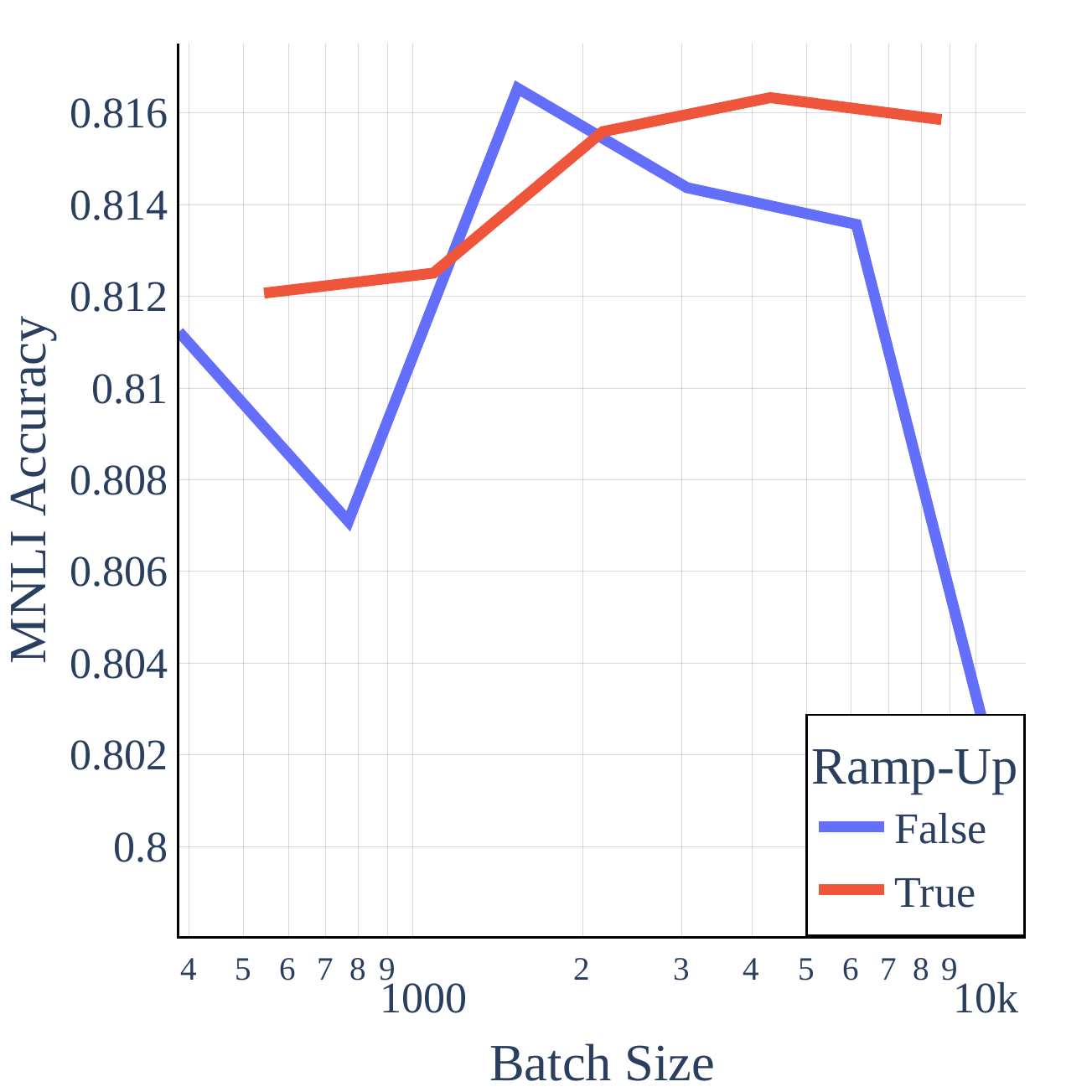}
    \caption{\looseness -1 Partial variations of batch sizes with and without linear ramp-up.  All experiments run with the training setup described in \cref{sec:training} for a day on a single GPU with mixed precision. Batch size is 4036 and dataset is \texttt{bookcorpus-wikipedia}. Downstream evaluation as described in \cref{sec:downstream}. All values for pretraining on an A4000. Note the discrepancy between optimal pretraining batch size and optimal batch size for evaluation on MNLI when ramp-up is used, but also note that differences are overall barely significant.}
    \label{fig:batch_szies}
\end{figure}

\subsection{References for Table 1}

The maximal floating point operations referenced in \cref{tab:flops} are based on the following published numbers. For TPU specs, according to \url{https://cloud.google.com/tpu/docs/system-architecture-tpu-vm} we find 275 TFLOP/s in \texttt{bfloat16} precision for the TPUv4 and 123 TFLOP/s for the TPUv3, each per chip. The V100 peak performance is given as 125 TFLOP/s in \url{https://images.nvidia.com/content/volta-architecture/pdf/volta-architecture-whitepaper.pdf} in "TFLOPS of mixed precision". Some NVIDIA datasheets also reference TFLOP/s with sparsity, which are not applicable in the context of this work. The Titan RTX comes out at 130.5 TFLOP/s, in "Peak FP16 Tensor TFLOPS with FP32
Accumulate" as described in \url{https://images.nvidia.com/aem-dam/en-zz/Solutions/geforce/ampere/pdf/NVIDIA-ampere-GA102-GPU-Architecture-Whitepaper-V1.pdf}. For the A6000, we find 154.8 "Peak BF16 Tensor TFLOPS with FP32 Accumulate" also in \url{https://www.nvidia.com/content/PDF/nvidia-ampere-ga-102-gpu-architecture-whitepaper-v2.pdf}. A4000 performance is actually less clear from this whitepaper, and estimated to be 88.45 TFLOP/s, based on it containing 192 tensor cores, compared to 336 for the A6000. For the RTX2080ti, the whitepaper at \url{https://images.nvidia.com/aem-dam/en-zz/Solutions/design-visualization/technologies/turing-architecture/NVIDIA-Turing-Architecture-Whitepaper.pdf} reports 53.8 "peak FP16 Tensor TFLOPS with FP32 Accumulate" for the reference edition. All total exaFLOP numbers are then computed based on these TFLOP/s numbers over the training time period described in each work.

\begin{table}
    \centering
    \small
\begin{tabular}{lrrrrrrrr|r}
\toprule
 & MNLI & SST-2 & STSB & RTE & QNLI & QQP & MRPC & CoLA & GLUE \\
\midrule
\hline
& \multicolumn{9}{c}{Trained for 1 day on a \textbf{2080ti}}\\
\hline
crammed BERT & 82.8 / 83.4 & 91.5 & 83.1 & 54.0 & 89.0 & 87.2 & 86.2 & 47.2 & 78.3 \\
+with original data & 81.7 / 82.0 & 91.3 & 82.3 & 51.8 & 88.7 & 86.9 & 85.5 & \textbf{48.1} & 77.6 \\
+with original train & 50.4 / 50.7 & 81.1 & 12.2 & 50.9 & 58.2 & 66.2 & 73.8 & 8.9 & 50.3 \\
+with original arch. & 58.7 / 57.8 & 79.8 & 16.6 & 50.9 & 55.4 & 71.1 & 70.1 & 7.3 & 52.0 \\
\hline
+with minimal train mod. & 80.2 / 80.5 & 89.6 & 82.7 & \textbf{55.4} & 86.6 & 86.4 & 84.1 & 39.0 & 76.0 \\
+with minimal arch. mod. & 81.7 / 82.5 & 91.2 & 79.2 & 54.5 & 87.7 & 86.4 & 83.0 & 38.8 & 76.1 \\
\hline
& \multicolumn{9}{c}{Trained for 1 day on an \textbf{A4000}}\\
\hline
crammed BERT  & 83.0 / 83.2 & 91.6 & 84.8 & 54.7 & 88.5 & 86.9 & 86.4 & 43.7 & 78.1 \\
+with original data & 81.5 / 81.8 & 91.0 & 81.8 & 49.5 & 88.3 & 86.8 & 84.5 & 43.2 & 76.5 \\
+with original train & 50.2 / 50.8 & 80.8 & 12.8 & 49.8 & 59.0 & 66.3 & 73.7 & 7.7 & 50.1 \\
+with original arch. & 58.0 / 56.5 & 79.4 & 17.0 & 51.6 & 54.2 & 70.6 & 74.1 & 8.2 & 52.2 \\
\hline
+with minimal train mod. & 80.0 / 80.4 & 89.3 & 84.2 & 55.2 & 86.5 & 86.4 & 86.3 & 40.1 & 76.5 \\
+with minimal arch. mod. & 82.1 / 82.6 & 91.5 & 79.9 & 54.7 & 87.9 & 86.6 & 82.9 & 35.4 & 76.0 \\
\hline
& \multicolumn{9}{c}{Trained for 1 day on an \textbf{A6000}}\\
\hline
crammed BERT  & \textbf{83.9} / \textbf{84.1} & \textbf{92.2} & 84.6 & 53.8 & \textbf{89.5} & \textbf{87.3} & 87.5 & 44.5 & \textbf{78.6} \\
+ original data & 82.2 / 82.7 & 92.0 & 83.6 & 49.8 & \textbf{89.5} & 87.0 & 85.9 & 42.5 & 77.3 \\
+ original train & 50.0 / 50.4 & 80.7 & 13.7 & 52.0 & 59.8 & 65.1 & 73.2 & 7.2 & 50.2 \\
+ original arch. & 35.4 / 35.2 & 49.1 & - & 52.7 & 49.5 & 0.0 & 0.0 & 0.0 & 27.7 \\
\hline
+ minimal train mod. & 81.9 / 82.6 & 91.4 & \textbf{85.5} & 54.9 & 88.2 & 87.0 & \textbf{88.4} & 43.6 & 78.1 \\
+ minimal arch. mod. & 83.2 / 83.5 & 91.7 & 82.0 & 52.0 & 88.9 & 86.8 & 83.6 & 38.3 & 76.7 \\
\bottomrule
\end{tabular}
    \caption{Extension of \cref{tab:ablation_data_train_arch}, including results on the other GPU types.
    }
    \label{tab:ablation_data_train_arch2}
\end{table}

\begin{table}
    \centering
        \begin{tabular}{lrrrr}
        \toprule
        Name & MLM Loss & MNLI-m & MNLI-mm & Tokens/Second \\
        \midrule
        %
        %
        Modified Transformer & 1.89 & 81.02 & 81.35 & 50946 \\
        DeepNarrow (12 Layers) & 1.94 & 80.90 & 80.97 & 78396 \\
        DeepNarrow (24 Layers) & 1.98 & 80.78 & 81.14 & 41289 \\
        $E=128$ & 2.14 & 76.68 & 77.62 & 53267 \\
        FFN every 2 blocks & 1.93 & 80.43 & 80.97 & 64774 \\
        FFN every 3 blocks & 1.97 & 80.44 & 80.93 & 71634 \\
        FFN every 4 blocks & 2.00 & 80.03 & 79.67 & 73319 \\
        $H=512$ & 1.93 & 80.61 & 80.93 & 83718 \\
        $H=1024$ & 1.95 & 80.07 & 80.68 & 32004 \\
        4 Layers & 2.00 & 78.45 & 79.00 & 137127 \\
        6 Layers & 1.93 & 79.49 & 79.82 & 96156 \\
        8 Layers & 1.89 & 81.11 & 81.08 & 74248 \\
        10 Layers & 1.89 & 81.02 & 81.21 & 61431 \\
        16 Layers & 1.92 & 81.39 & 82.10 & 39406 \\
        24 Layers & 2.01 & 80.64 & 80.97 & 26927 \\
        Recurrent (1-12) & 2.40 & 77.46 & 77.81 & 52405 \\
        Recurrent (2-6) & 2.04 & 80.45 & 80.73 & 53148 \\
        Recurrent (3-4) & 2.00 & 80.78 & 81.33 & 51634 \\
        Recurrent (4-3) & 1.98 & 80.95 & 81.26 & 51952 \\
        \hline
        BERT-tiny & 3.30 & 56.71 & 57.21 & 914694 \\
        BERT-mini & 2.49 & 72.22 & 73.21  & 429593 \\
        BERT-Large (Izsak variant) & 2.38 & 76.93 & 77.47 & 13448 \\
        Original BERT & 7.54 & 35.45 & 35.22 & 41978 \\
        \hline
        With decoder bias & 1.89 & 80.97 & 81.20 &  51155 \\
        With $\varepsilon=\num{e-6}$ in Layer Norm & 1.90 & 80.49 & 81.35 & 51728 \\
        Learned Embedding & 1.88 & 80.51 & 81.03 & 52601 \\
        No Norm after Embedding & 1.94 & 79.65 & 80.34  & 52175 \\
        No Final Norm & 1.89 & 80.40 & 80.89 & 51207 \\
        No Skip of Head Transform & 1.88 & 80.49 & 81.19 & 51728 \\
        No Rotational Embedding & 1.88 & 80.91 & 81.52 & 53526 \\
        Post-LN & 7.54 & 31.82 & 31.82 & 52270 \\
        With QKV bias & 1.89 & 80.70 & 80.88 & 51112 \\
        With bias in Linear Layers & 1.89 & 80.64 & 81.49 & 50584 \\
        12 Heads & 1.88 & 81.75 & 81.99 & 47967 \\

        \bottomrule
        \end{tabular}
    \caption{Additional raw results for experiments considered in the main body. This table contains architecture variants for a prelimary architecture setup which contained 4 heads in the attention block, 12 layers and included rotary embeddings. First two blocks: Architectural variants as discussed in \cref{sec:arch} (but for this preliminary variant). Third block: Ablation study of this model. All experiments run with the training setup described in \cref{sec:training} for a day on a single GPU with mixed precision. Batch size is 4032 and dataset is \texttt{bookcorpus-wikipedia}. Downstream evaluation as described in \cref{sec:downstream}. All values for pretraining on an A4000.}
    \label{tab:arch1}
\end{table}

\begin{table}
    \centering
\begin{tabular}{lrrrr}
\toprule
Name & MLM Loss & MNLI & MNLI-mm & Tokens/Second \\
\midrule
Modified Transformer & 1.84 & 81.79 & 82.14 & 46431 \\
DeepNarrow (12 Layers) & 1.91 & 80.97 & 81.30 & 99717 \\
DeepNarrow (24 Layers) & 1.91 & 81.39 & 81.61 & 52558 \\
$E=128$ & 2.04 & - & - & 48468 \\
FFN every 2 blocks & 1.84 & 81.40 & 81.65 & 62134 \\
FFN every 3 blocks & 1.87 & 80.90 & 81.53 & 70685 \\
FFN every 4 blocks & 1.88 & 81.10 & 81.42 & 75163 \\
$H=512$ & 1.87 & 81.34 & 82.20 & 79116 \\
$H=1024$ & 1.94 & 80.63 & 80.97 & 28511 \\
4 Layers & 1.94 & 79.13 & 79.51 & 161034 \\
6 Layers & 1.87 & 80.48 & 80.84 & 115037 \\
8 Layers & 1.84 & 81.22 & 81.62 & 88652 \\
10 Layers & 1.82 & 81.25 & 82.31 & 71414 \\
12 Layers & 1.85 & 81.68 & 82.18 & 59346 \\
18 Layers & 1.90 & 81.02 & 81.82 & 40577 \\
24 Layers & 1.97 & 80.81 & 81.26 & 30455 \\

Recurrent (1-12) & 2.13 & 79.23 & 79.78 & 62318 \\
Recurrent (2-6) & 2.00 & 80.86 & 81.24 & 62677 \\
Recurrent (3-4) & 1.94 & 80.95 & 81.48 & 61772 \\
Recurrent (4-3) & 1.91 & 81.43 & 81.84 & 61596 \\
\hline
BERT-Tiny Variant & 3.51 & 56.10 & 56.60 & 1018443 \\
BERT-Mini Variant & 2.46 & 72.30 & 73.47 & 523061 \\
BERT-Large Variant & 2.12 & 79.50 & 79.84 & 17688 \\
BERT-Large (Izsak variant) & 2.37 & 76.81 & 77.56 & 13522 \\
Original BERT & 7.53 & 35.45 & 35.22 & 41362 \\
\hline
With decoder bias & 1.84 & 81.71 & 81.91 & 45996 \\
With $\varepsilon=\num{e-6}$ in Layer Norm & 1.83 & 81.55 & 82.13 & 45841 \\
Learned Embedding & 1.83 & 81.31 & 81.79 & 46608 \\
No Norm after Embedding & 1.89 & 81.38 & 81.15 & 46267 \\
No Final Norm & 1.85 & 80.67 & 80.87 & 46598 \\
No Skip of Head Transform & 1.83 & 82.03 & 82.19 & 46324 \\
With QKV Bias & 1.83 & 81.89 & 82.28 & 46469 \\
With bias in Linear Layers & 1.84 & 81.88 & 82.16 & 45629 \\
4 Heads & 1.88 & 81.22 & 81.77 & 40551 \\
With Rotary Embedding & 1.86 & 81.16 & 81.94 & 42257 \\
Post-LN & 7.54 & 35.21 & 35.17 & 46324 \\
Fourier Attention & 2.65 & 68.97 & 69.06 & 46634 \\
GELU & 1.832477 & 81.94 & 82.17 & 47779 \\
\bottomrule
\end{tabular}
    \caption{Additional raw results for experiments considered in the main body for the final architecture variant. First two blocks: Architectural variants as discussed in \cref{sec:arch}. Third block: Ablation study of finally adopted model. All experiments run with the training setup described in \cref{sec:training} for a day on a single GPU with mixed precision. Batch size is 4096 and dataset is \texttt{bookcorpus-wikipedia}. Downstream evaluation as described in \cref{sec:downstream}. All values for pretraining on an A4000.}
    \label{tab:arch2}
\end{table}

\begin{table}
    \centering
\begin{tabular}{lrrrr}
\toprule
Name & MLM & MNLi-m & MNLI-mm & Tokens/Second \\
\midrule
Original training recipe & 7.28 & 60.65 & 60.31 & 49264 \\
With Izsak Training recipe & 2.06 & 79.90 & 80.30 & 46869 \\
Minimal Modifications & 2.03 & 78.78 & 79.36 & 47346 \\
+Larger LR & 1.99 & 80.25 & 80.50 & 46524 \\
+One Cycle, +Larger LR & 1.84 & 82.12 & 82.55 & 46843 \\
+One Cycle, +Larger LR, +Clipping & 1.84 & 81.79 & 82.14 & 46303 \\
\hline
Sequence Curriculum (10\%,20\%,30\%,50\%,75\%) & 3.02 & 70.06 & 70.77 & 29359 \\
Sequence Curriculum (+unfolding) & 1.87 & 80.13 & 80.04 & 46014 \\
Sequence Curriculum (20\%,35\%,50\%,65\%,85\%) & 1.90 & 79.86 & 79.80 & 45804 \\
Adafactor & 1.86 & 81.36 & 82.22 & 45997 \\
Adam (classic WD formulation) & 7.44 & 32.28 & 32.39 & 49598 \\
SGD & 7.46 & 59.30 & 58.02 & 47678 \\
RADAM & 7.50 & 32.74 & 32.95 & 48812 \\
With Dropout activated & 1.97 & 80.95 & 80.98 & 45198 \\
\hline
With MLM masking 20\% & 2.06 & 80.76 & 81.48 & 45944 \\
With MLM masking 40\% & 2.70 & 81.11 & 81.30 & 43467 \\
With MLM masking 60\% & 3.41 & 80.62 & 80.88 & 40756 \\
\bottomrule
\end{tabular}
    \caption{Additional raw results for experiments considered in the main body for the final training variant, not otherwise mentioned. Batch size is 4096 and dataset is \texttt{bookcorpus-wikipedia}. Downstream evaluation as described in \cref{sec:downstream}. All values for pretraining on an A4000.}
    \label{tab:train_raw}
\end{table}

\end{document}